\documentclass[10pt,journal,compsoc]{IEEEtran}
\ifCLASSOPTIONcompsoc
  \usepackage[nocompress]{cite}
\else
  \usepackage{cite}
\fi

\ifCLASSINFOpdf
\else
\fi

\usepackage{graphicx}
\usepackage{amsmath}
\usepackage{amssymb}
\usepackage{booktabs}
\usepackage{tabulary}
\usepackage{multirow}
\usepackage{overpic}
\usepackage{xspace}
\usepackage{epsfig}
\usepackage{hyperref}
\usepackage{graphicx}
\graphicspath{ {./figures/} }
\usepackage{multicol}
\usepackage{epsfig}
\usepackage{hyperref}
\usepackage{graphicx}
\usepackage{amsmath}
\usepackage{amssymb}
\usepackage{booktabs}
\usepackage{tabulary}
\usepackage{multirow}
\usepackage{overpic}
\usepackage{xspace}
\usepackage[ruled]{algorithm2e}
\usepackage{algorithmic}
\usepackage{bbm}
\newcommand{\vct}[1]{\boldsymbol{#1}} 
\newcommand{\mat}[1]{\boldsymbol{#1}} 

\newcommand{\methodname}{{MTLFace}\xspace}
\newcommand{\seldaname}{{FT-Sel}\xspace}
\newcommand{\testsetname}{{ECAF}\xspace}

\newcommand{\app}{\raise.17ex\hbox{$\scriptstyle\sim$}}

\newcommand{\etal}{\textit{et al}.\xspace}
\newcommand{\ie}{\textit{i}.\textit{e}.\xspace}
\newcommand{\eg}{\textit{e}.\textit{g}.\xspace}
\newcommand{\std}[1]{{{\scriptsize $\pm$#1}}} 

\hypersetup{
          breaklinks=true,   
          colorlinks=true,   
          pdfusetitle=true,  
      }

\begin{document}

\title{When Age-Invariant Face Recognition Meets Face Age Synthesis: A Multi-Task Learning Framework and A New Benchmark}

\author{Zhizhong~Huang,
Junping~Zhang
and~Hongming~Shan
\IEEEcompsocitemizethanks{\IEEEcompsocthanksitem Z. Huang and J. Zhang are with the Shanghai Key Lab of Intelligent Information Processing and the School of Computer Science, Fudan University, Shanghai 200433, China.\protect\\ 
Email: \{zzhuang19,jpzhang\}@fudan.edu.cn.
\IEEEcompsocthanksitem H. Shan is with the Institute of Science and Technology for Brain-inspired Intelligence and  MOE Frontiers Center for Brain Science and Key Laboratory of Computational Neuroscience and Brain-Inspired Intelligence (Ministry of Education), Fudan University, Shanghai, 200433, China, and also with the Shanghai Center for Brain Science and Brain-inspired Technology, Shanghai 201210, China.\protect\\ 
E-mail: hmshan@fudan.edu.cn.}
\thanks{(Corresponding author: Hongming Shan)}
}


\IEEEtitleabstractindextext{%

\begin{abstract}

To minimize the impact of age variation on face recognition, age-invariant face recognition (AIFR) extracts identity-related discriminative features by minimizing the correlation between identity- and age-related features while face age synthesis (FAS) eliminates age variation by converting the faces in different age groups to the same group. However, AIFR lacks visual results for model interpretation and FAS compromises downstream recognition due to artifacts. 
Therefore, we propose a unified, multi-task framework to jointly handle these two tasks, termed \methodname, which can learn the age-invariant identity-related representation for face recognition while achieving pleasing face synthesis for model interpretation.
Specifically, we propose an attention-based feature decomposition to decompose the mixed face features into two uncorrelated components---identity- and age-related features---in a spatially constrained way. 
Unlike the conventional one-hot encoding that achieves group-level FAS, we propose a novel identity conditional module to achieve identity-level FAS, which can improve the age smoothness of synthesized faces through a weight-sharing strategy.
Benefiting from the proposed multi-task framework, we then leverage those high-quality synthesized faces from FAS to further boost AIFR via a novel  selective fine-tuning strategy. Furthermore, to advance both AIFR and FAS, we collect and release a large cross-age face dataset with age and gender annotations, and a new benchmark specifically designed for tracing long-missing children.
Extensive experimental results on five benchmark cross-age datasets demonstrate that \methodname yields superior performance than state-of-the-art methods for both AIFR and FAS.
We further validate \methodname on two popular general face recognition datasets, obtaining competitive performance on face recognition in the wild. The source code and datasets are available at~\url{http://hzzone.github.io/MTLFace}.
\end{abstract}

}

\maketitle

\IEEEdisplaynontitleabstractindextext

\IEEEpeerreviewmaketitle

\bstctlcite{IEEEexample:BSTcontrol}

\section{Introduction}
\begin{figure}[t]
    \centering
    \includegraphics[width=1\linewidth]{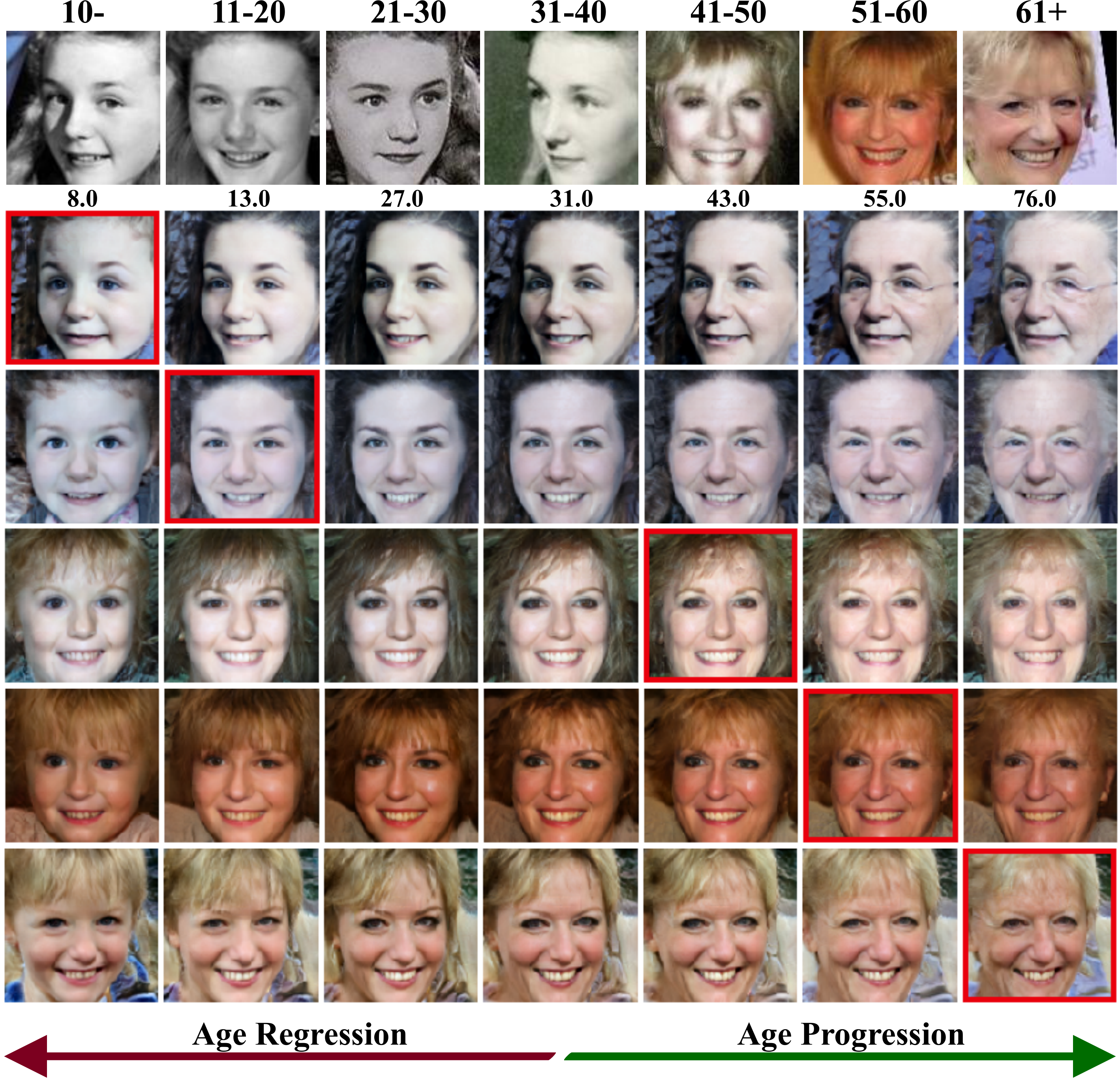}
    \caption{Sample results of our \methodname. First row: real faces of the same person at different ages with estimated age labels underneath. Remaining rows: synthesized faces where the input faces are given in the red boxes.}
    \label{fig:example}
\end{figure}

\IEEEPARstart{F}{ace} recognition has been a hot research topic in computer vision for many years. Recently, deep-learning-based methods have achieved excellent performance, even surpassing humans in several scenarios, by empowering face recognition models with deep neural networks~\cite{he2016deep,krizhevsky2017imagenet,simonyan2014very}. The traditional wisdom is to train the face recognition models to increase the intra-class compactness with a massive amount of data and margin-based metrics for improved recognition performance~\cite{wen2016discriminative,meng2021magface,meng2021learning}.

Despite the remarkable success of general face recognition (GFR), \emph{how to minimize the impact of age variation} is a lingering challenge for current face recognition systems in correctly identifying faces in many practical applications such as tracing long-missing children. Therefore, it is of great significance to achieve face recognition without age variation, \ie, age-invariant face recognition or AIFR. AIFR, however, remains extremely challenging in the following three aspects. First, when the age gap becomes large in cross-age face recognition, age variation can dominate the facial appearance, which then significantly compromises the face recognition performance. Second, face age synthesis (FAS) is a complex process involving face aging/rejuvenation~(\emph{a.k.a} age progression/regression) since the facial appearance changes dramatically over a long time and differs from person to person. Last, it is infeasible to obtain a large-scale paired face dataset for training a model to render faces with natural effects while preserving identities.

To address the issues mentioned above, current methods for AIFR can be roughly summarized into two categories: generative and discriminative models. Given a face image, generative models~\cite{geng2007automatic,lanitis2002toward,park2010age} aim to transform faces of different age groups into the same age group in order to minimize the impact of age variation on face recognition. Recently, generative adversarial networks~(GANs)~\cite{goodfellow2014generative} have been successfully used to enhance the image quality of synthesized faces~\cite{li2019age,liu2019attribute,wang2018face,yang2018learning,zhang2017age}, which typically use one-hot encoding to specify the target age group. However, the one-hot encoding represents the age group-level face transformation, ignoring identity-level personalized patterns and leading to unexpected artifacts. As a result, the performance of AIFR cannot be significantly improved due to the unpleasing synthesized faces and unexpected change in identity. On the other hand, discriminative models~\cite{deb2019finding,wang2018orthogonal} focus on extracting age-invariant features by disentangling the identity-related information from mixed face information so that only the identity-related information is used by the face recognition systems. Although these models achieve promising performance in AIFR, they cannot provide users, for example policemen, with visual results just as the generative methods do to further verify identities, which can compromise model interpretability in the decision-making processes of many practical applications.

To further improve the image quality of generative models and provide model interpretability for discriminative models, in this paper, we propose a unified, multi-task learning framework to simultaneously achieve AIFR and FAS, termed \methodname, which can enjoy the best of both worlds; \ie, learning age-invariant identity-related representations while achieving pleasing face synthesis. More specifically, we propose an attention-based feature decomposition to decompose the mixed high-level features into two uncorrelated components---identity- and age-related features---in a spatially constrained way. We then decorrelate these two components in a multi-task learning framework, in which an age estimation task is to extract age-related features while a face recognition task is to extract identity-related features; in addition, a continuous cross-age discriminator with a gradient reversal layer~\cite{ganin2016domain} further encourages extracting identity-related age-invariant features.
Moreover, we propose an identity conditional module to achieve identity-level transformation patterns for FAS, with a weight-sharing strategy to improve the age smoothness of the synthesized faces; \ie, the faces are aged smoothly.
Extensive experimental results demonstrate the superior performance of the proposed \methodname over existing state-of-the-art methods for AIFR and FAS, and competitive performance for general face recognition in the wild. Fig.~\ref{fig:example} presents an example of age progression and regression of the same person by our \methodname, showing that our framework can synthesize photorealistic faces while preserving identity. 

The contributions of this paper are summarized as follows.
\begin{enumerate} 
\item We propose a unified, multi-task learning framework to jointly handle AIFR and FAS, which can learn age-invariant identity-related representations while achieving pleasing face synthesis. 
\item We propose an attention-based feature decomposition to separate the age- and identity-related features on high-level feature maps, which can spatially constrain the decomposition process in contrast to the previous unconstrained decomposition on feature vectors. Age estimation and face recognition tasks are incorporated to supervise the decomposition process in conjunction with a continuous domain adaptation. 
\item We propose a novel identity conditional module to achieve identity-level face transformation, which leverages a weight-sharing strategy to improve the age smoothness of the synthesized faces in contrast to previous one-hot encoding that can only achieve age group-level face transformation. We further extend it into a multi-level architecture for improved face age synthesis.
\item We propose a selective fine-tuning strategy to further boost AIFR by automatically selecting those high-quality synthesized faces from FAS for fine-tuning.
\item We collect and release a large-scale cross-age dataset of millions of faces with balanced age and gender annotations, which can not only advance the development of the AIFR and FAS but also be useful for other face-related research tasks; \eg, pretraining for face age estimation. 
\item To promote the utility in tracing long-missing children, we construct a new benchmark with paired child and adult faces of the same person, which contains the same identities as the Labeled Faces in the Wild (LFW) and is thus orthogonal to the current literature for future evaluation of cross-age face recognition.
\item Extensive experimental results demonstrate the effectiveness of the proposed framework for AIFR and FAS on five benchmark datasets, and competitive performance on two popular GFR datasets. 
\end{enumerate}

We note that a preliminary version of this work was published in IEEE/CVF Conference on Computer Vision and Pattern Recognition (CVPR) 2021~\cite{huang2021age}. In this paper, we further extend our preliminary work with the following major improvements. 
First, we propose a selective fine-tuning strategy that selects the high-quality synthesized faces from FAS through a two-component Gaussian Mixture model and then fine-tunes the model to further boost AIFR, bringing consistent performance improvements on all benchmark datasets.
Second, we extend the model by incorporating multi-level age conditional information with a StyleGAN-based architecture, which, as a result, can synthesize more photorealistic faces than previous model~\cite{huang2021age} and achieve continuous face age synthesis.
Third, we construct a new benchmark testing set for the AIFR task with an emphasis on tracing long-missing children, which contains a subset of identities of LFW with manually annotated and paired child-adult faces; extensive baselines including humans and other deep-learning based face recognition models were conducted to evaluate the new benchmark.
Fourth, we further conduct more experiments with more competitive baselines to evaluate the face age synthesis task of our model including the results on the new benchmark.
Finally, we evaluate the model through both quantitative and qualitative comparisons between aged/rejuvenated faces and ground-truth faces to further validate the effectiveness of the proposed method. We also visualize the training process of \methodname to show its training stability. Note that a detailed comparison between the current and preliminary versions of \methodname can be found in supplementary Sec.~\ref{sec:comparisons_to_conf}; our new model consistently improves the performance of face recognition and the quality of synthesized faces.

The remainder of this paper is organized as follows. A brief review of the related work on age-invariant face recognition and face age synthesis is given in Section~\ref{sec:related_work}. We present the attention-based feature decomposition, identity conditional module, and selective fine-tuning strategy as well as
the proposed \methodname in Section~\ref{sec:method}. Comprehensive experimental results on the age-invariant face recognition and face age synthesis are reported and analyzed in
Section~\ref{sec:exp}. Finally, this paper is discussed in Section~\ref{sec:discussion}, followed by a concluding summary in Section~\ref{sec:conc}. 

\begin{figure*}[ht!]
    \centering
    \includegraphics[width=1\linewidth]{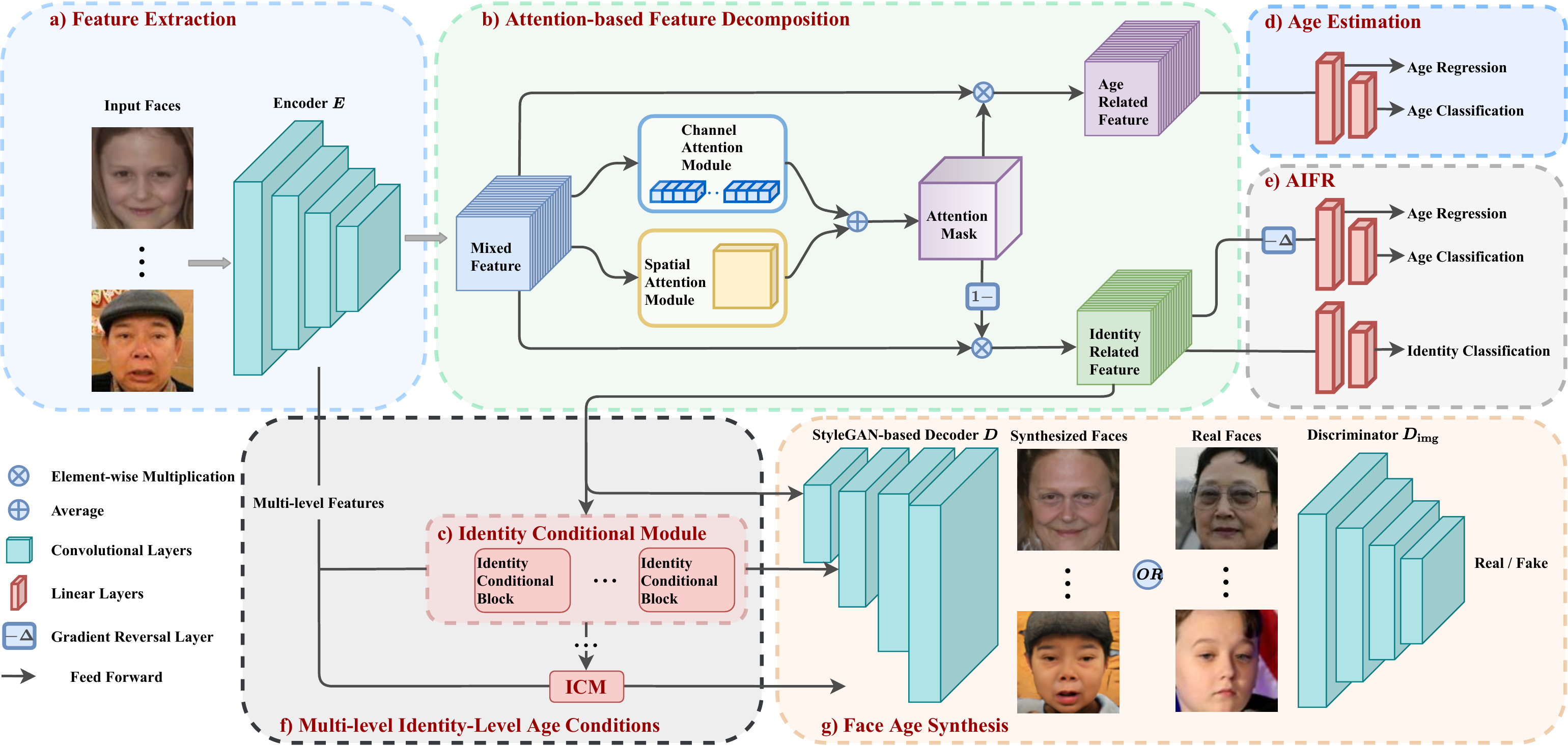}
    \caption{An overview of the proposed \methodname including two tasks. AIFR: The encoder $E$ first extracts the mixed feature maps from input faces, which are then decomposed into two disjoint identity- and age-related feature maps by the multi-task training and continuous domain adaptation. FAS: The decoder $D$ produces synthesized faces from identity-level features under the control of multi-level age conditions from the output of the identity conditional modules; the discriminator $D_{\mathrm{img}}$ penalizes the framework to obtain better visual quality. 
    }
    \label{fig:framework}
\end{figure*}
\section{Related Work}
\label{sec:related_work}

This section briefly surveys developments in age-invariant face recognition and face age synthesis.

\subsection{Age-invariant Face Recognition (AIFR)}

Prior studies usually minimize the impacts of age variation by disentangling age-invariant features from mixed features. For example, Gong~\etal adopted hidden factor analysis~(HFA) to factorize mixed features and then minimize the age variation in identity-related features~\cite{gong2013hidden}. Later, Wen~\etal extended HFA to a deep learning framework with the latent factor guided convolutional neural network~(LF-CNN)~\cite{wen2016latent}. Zheng~\etal introduced an age estimation task to guide AIFR~\cite{zheng2017age}. Most recently, CNNs-based discriminative methods have achieved promising results for AIFR. To reduce the intra-class discrepancy caused by aging, Wang~\etal proposed an orthogonal embedding CNN for AIFR~\cite{wang2018orthogonal}, termed OE-CNN, which decomposes the facial embeddings into two orthogonal components, where the identity- and age-related features are represented as the angular and radial directions, respectively.
Similarly, Wang~\etal proposed the decorrelated adversarial learning~(DAL) to achieve feature decomposition in an adversarial manner under the assumption that the two components are uncorrelated~\cite{wang2019decorrelated}.
Lee~\etal proposed an inter-prototype loss to minimize the similarity between child faces~\cite{lee2021improving}. Hou~\etal~\cite{hou2021disentangled} and Xie~\etal~\cite{xie2022implicit} proposed to minimize the mutual information between the identity- and age-related components of the face image from the same person to reduce the effect of age variations.

The work related to ours is~\cite{zhao2019look}, in which a cGANs-based model with cross-age domain adversarial training to extract age-invariant representations, is adopted to perform the two tasks simultaneously. However, it generates oversmoothed faces with subtle changes. Different from~\cite{zhao2019look}, our framework has the following advantages:
1) our feature decomposition is performed on feature maps through an attention mechanism; 
2) a continuous domain adaptation with a gradient reversal layer is used to learn age-invariant identity-related representation; 
3) the proposed identity conditional module can achieve identity-level face synthesis and improve the age smoothness of synthesized faces; and
4) the proposed \seldaname can further boost the performance of AIFR with the automatically selected high-quality synthesized faces from FAS.

\subsection{Face Age Synthesis (FAS)}

Existing methods for FAS can be roughly divided into physical model-, prototype-, and deep generative model-based methods. 

Physical model-based methods~\cite{ramanathan2006modeling,ramanathan2008modeling,suo2012concatenational} mechanically model the changes in faces over time, but they are computationally expensive and require massive paired images of the same person over a long period of time. Prototype-based methods~\cite{kemelmacher2014illumination,rowland1995manipulating} achieve face aging/rejuvenation using the averaged faces in each age group, and consequently, the identity cannot be well preserved. 

In contrast to the above two categories, deep generative model-based methods~\cite{Duong_2017_ICCV,wang2016recurrent} exploit a deep neural network for this task. For example, Wang~\etal proposed recurrent face aging~(RFA), which uses a recurrent neural network to model the intermediate transition states of age progression/regression, traversing on which a smooth face aging process can be achieved~\cite{wang2016recurrent}. Inspired by the powerful capability of generative adversarial networks~(GANs)~\cite{goodfellow2014generative}, especially conditional GANs~(cGANs)~\cite{mirza2014conditional}, in generating high-quality images, many recent studies~\cite{despois2020agingmapgan,huang2021ageflow,lihierarchical,or2020lifespan,huang2020pfa,zhang2017age,wang2018face,yang2018learning} resort to them to improve the visual quality of synthesized faces and train models with unpaired face images. For example, Zhang~\etal used a conditional adversarial autoencoder~(CAAE) to achieve both age progression/regression by traversing on a low-dimensional face manifold~\cite{zhang2017age}. Wang~\etal introduced the perceptual loss to preserve identities during face aging/rejuvenation~\cite{wang2018face}. Yang~\etal designed a discriminator with a pyramid architecture to enhance aging details~\cite{yang2018learning,yang2019learning}.

However, these methods mainly aim to improve the visual quality of generated faces, and hardly improve the performance of AIFR due to the artifacts resulting from group-level face transformation, and the unexpected change in identity. Our method differs in the following aspects: 
1) the proposed \methodname achieves AIFR and FAS simultaneously to enhance the visual quality with identity-related information from AIFR; 
2) the proposed identity conditional module~(ICM) achieves an identity-level face age synthesis in contrast to the previous group-level face age synthesis;
3) a weight-sharing strategy in ICM can improve the age smoothness of the synthesized faces; and
4) a well-designed generator with feature-pyramid and StyleGAN architectures is employed for photorealistic and continuous synthesis.
\section{Methodology}\label{sec:method}

Fig.~\ref{fig:framework} presents the architecture of the proposed \methodname, which is detailed in the following subsections. 

\subsection{Attention-based Feature Decomposition}

As the faces age over time, the critical problem of AIFR is that the age variation usually increases intra-class distances. As a result, it becomes challenging to correctly recognize two face images of the same person with a large age gap, since the mixed facial representations are severely entangled with unrelated information such as facial shape and texture changes. Recently, Wang~\etal~designed a linear factorization module to decompose feature vectors into two unrelated components~\cite{wang2019decorrelated}. Formally, the feature vector $\vct{x}\in\mathbb{R}^d$, extracted from an input image $\mat{I}\in \mathbb{R}^{3\times H\times W}$, can be decomposed as~\cite{wang2019decorrelated}:
\begin{align}\label{eq_linear_decomposition}
    \mat{x} = \mat{x}_{\mathrm{age}} + \mat{x}_{\mathrm{id}},
\end{align}
where $\mat{x}_{\mathrm{age}}$ and $\mat{x}_{\mathrm{id}}$ denote the age- and identity-related components, respectively. This decomposition is implemented through a residual mapping. However, it has the following drawbacks: 1) this decomposition performs on one-dimensional feature vector, and the resultant identity-related component lacks spatial information of face, not suitable for FAS; and 2) this decomposition is unconstrained, which may lead to unstable training. 

To address these drawbacks, we instead propose to decompose the mixed feature-maps in a high-level semantic space through an attention mechanism, termed attention-based feature decomposition or AFD. The main reason is that manipulating on the feature vectors is more complicated than on the feature maps since the aging/rejuvenation effects, such as beards and wrinkles, appear in the semantic feature space but are lost in one-dimensional feature vectors.  Formally, we use a ResNet-like backbone as encoder $E$ to extract mixed feature maps  $\mat{X} \in \mathbb{R}^{C\times H'\times W'}$ from an input image $\mat{I}$, \ie $\mat{X} = E(\mat{I})$, the AFD can be expressed as:
\begin{align}
\mat{X} = \underbrace{\mat{X} \circ \sigma(\mat{X})}_{\mat{X}_{\mathrm{age}}} + \underbrace{\mat{X} \circ \big(1 - \sigma(\mat{X})\big)}_{\mat{X}_{\mathrm{id}}},
\end{align}
where $\circ$ denotes element-wise multiplication and $\sigma$ represents an attention module. In doing so, the age-related information in the feature maps can be separated through the attention module supervised by an age estimation task, and the residual part, regarded as the identity-related information, can be supervised by a face recognition task. Consequently, the attention mechanism constrains the decomposition module, better at detecting the age-related features in semantic feature maps. We note that $\mat{X}$ is assumed to only contain the age and identity information as driven by the two corresponding tasks, and the remaining information such as background is important for FAS, which is preserved by skip connections from encoder to decoder. Fig.~\ref{fig:framework}(b) details the proposed AFD. 

In this paper, we adopt the average of channel attention~(CA)~\cite{hu2018squeeze} and spatial attention~(SA)~\cite{woo2018cbam} to highlight age-related information at both channel and spatial levels. Note that the outputs of these two attentions have different sizes; we first stretch each of them to the original input size and then average them. Different attention modules such as CA, SA, and CBAM~\cite{woo2018cbam}, are also investigated in Sec.~\ref{sec:exp}.

\subsection{Identity Conditional Module}

\begin{figure}[t]
    \centering
    \includegraphics[width=1\linewidth]{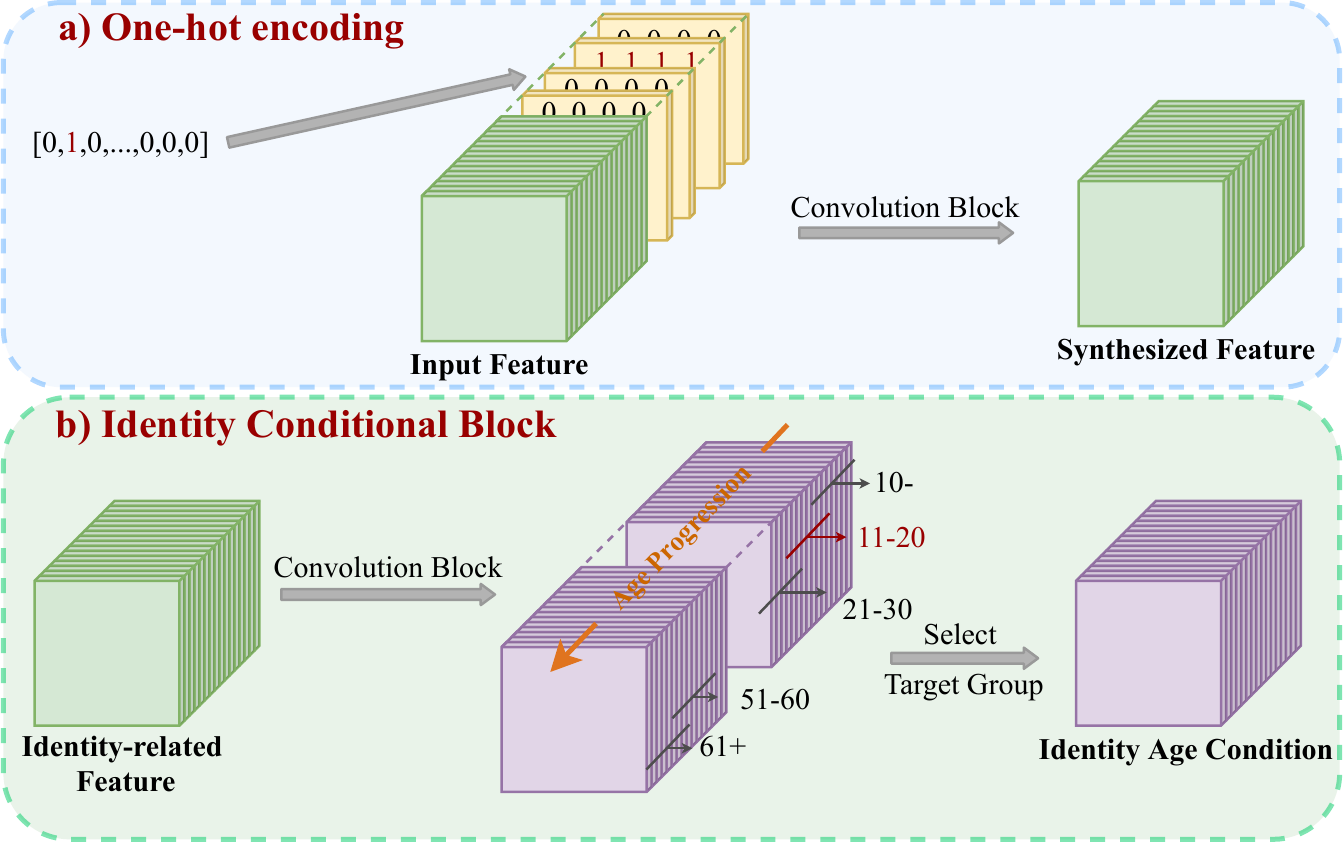}
    \caption{Comparison between one-hot encoding and ICB.
    }
    \label{fig:condition}
\end{figure}

The mainstream face aging studies~\cite{li2019age,liu2019attribute,wang2018face,yang2018learning,zhang2017age} usually split ages into several non-overlapping age groups, since the changes over time are minor with a small age gap. These methods typically use one-hot encoding to specify the target age group to control the aging/rejuvenation process~\cite{li2019age,wang2018face,zhang2017age} as illustrated in Fig.~\ref{fig:condition}(a). Consequently, a group-level aging/rejuvenation pattern, such as having a beard at age 30, is learned for each age group due to the use of one-hot age conditions. This has two drawbacks: 1) the one-hot encoding represents the age group-level aging/rejuvenation pattern, ignoring identity-level personalized patterns, particularly for different genders and races; and 2) one-hot encoding may not ensure the age smoothness of synthesized faces.

To address these issues raised by one-hot encoding, we propose an identity conditional block (ICB) to achieve identity-level aging/rejuvenation pattern, with a weight-sharing strategy to improve the age smoothness of synthesized faces.
Specifically, the proposed ICB takes the identity-related feature from AFD as input to learn an identity-level aging/rejuvenation pattern.
Note that the input features for ICB may be mixed up with the ages due to the use of unpaired training data, which could harm the learning of identity-level pattern. Instead, in our multi-task learning framework, under the supervision from face classification and domain adaptation loss, the identity features are invariant to ages. As a result, the ages would not have an impact on the training of ICB.
Next, we propose a weights-sharing strategy to improve the age smoothness of synthesized faces so that some convolutional filters are shared across adjacent age groups as shown in Fig.~\ref{fig:condition}(b).
The rationale behind this idea is that faces gradually change over time, and the shared filters can learn some common aging/rejuvenation patterns between adjacent age groups. 
Note that the number of channels of the input feature maps is reduced to $1/4$ of the original ones using $1\times 1$ convolutions to reduce the computational cost.
In this paper, a hyper-parameter $s$ is used control how many filters are shared for two adjacent age groups, which is empirically set to $1/8$; \ie, two adjacent age groups share 16 filters. We stack ICBs to form an identity conditional module (ICM), as shown in Fig.~\ref{fig:framework}(c). 

\subsection{Selective Fine-tuning Strategy}
\label{sec:ftsel}
In this section, we present a novel selective fine-tuning strategy~(\seldaname), which can \emph{selectively} include high-quality synthesized faces into the training data to further boost the performance of face recognition for AIFR task.

Our motivation comes from the observation that there are a few child faces below 10 years old in the training data since it is cumbersome to collect the paired child and adult faces. Such imbalanced data (\ie lacking child faces) would significantly harm the face recognition performance on child's faces, affecting the practical application for tracking long-missing children. As a result,  the intra-class distances between children are not discriminative, which has been widely reported in long-tailed learning~\cite{zhang2021deep}.

\begin{figure}[t]
    \centering
    \includegraphics[width=1\linewidth]{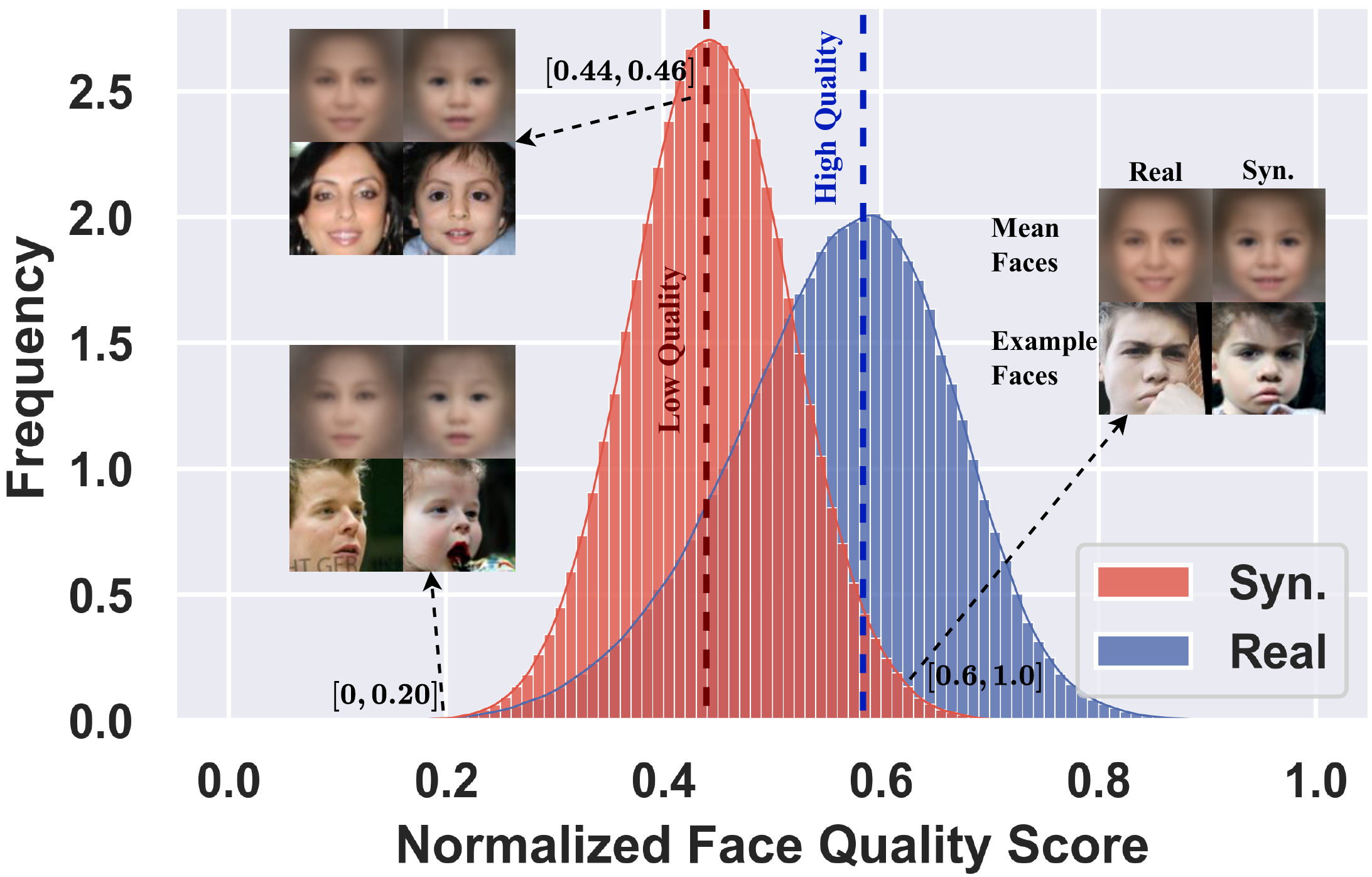}
    \caption{
        Visualization of the normalized face quality scores of synthesized and real faces,
        where the two vertical dotted lines denote the mean values of two components of the GMM, and
        a higher value indicates better quality. We showcase some examples with their rejuvenated faces, and the mean faces at corresponding scores. 
        }
    \label{fig:syn_data_augmentation}
\end{figure}

Benefiting from the proposed multi-task framework, we address this problem by transforming the faces above 10 years old into 10- to obtain the paired child and adult faces. 
This is similar to~\cite{hong2021stylemix} that leverages style transfer~\cite{gatys2016image} to augment the training data. If \methodname disentangles the faces to identity~(\ie content) and age~(\ie style), we are able to alter the age styles at low-level visual images while preserving the semantic identities. As a result, the face recognition model can be encouraged to learn style-invariant identity features to further boost performance.
However, only high-quality synthesized faces should be considered, as the ones with unexpected artifacts would harm the performance of face recognition~\cite{qiu2021synface}. To this end, we propose a selective fine-tuning strategy~(\seldaname), which can automatically select the high-quality synthesized faces using the face quality scores. Fig.~\ref{fig:syn_data_augmentation} visualizes the face quality scores~\cite{meng2021magface} of synthesized and real faces, where a significant discrepancy between the two kinds of faces can be observed. The synthesized faces with the low face quality scores~(\eg [0, 0.2]) present strong artifacts. When the scores increase, the synthesized faces become more photorealistic despite some occlusion~(\eg [0.6,1.0]). The corresponding mean face reveals more details at the same time. This implies that the face quality scores can measure the quality of synthesized faces. As a result, the synthesized child faces of better face quality can be selected to construct more balanced training data for further boosting the performance of face recognition.

A natural question arises: \emph{how to select those high-quality faces?} A straightforward way is to set a threshold, where the faces above this threshold can be considered as high-quality samples. However, the introduced thresholding hyper-parameter needs to be \emph{manually} tuned for satisfied recognition performance. By visualizing the distributions of the scores in Fig.~\ref{fig:syn_data_augmentation}, we propose to model the distribution of face quality scores of  synthesized and real faces by a two-component Gaussian Mixture Model (GMM) for \emph{automatically} selecting high-quality faces, which is defined as follows:
\begin{align}
    p(s)=\sum_{c=0}^{1}p(s, c) = \sum_{c=0}^{1} p(c)p(s | c),\label{eq:gmm_two}
\end{align}
where $s$ denotes the face quality score, and $c$ is a latent variable:
the cluster $c=1$ with a higher mean face-quality score denotes the faces with higher quality while $c=0$ corresponds to the faces with lower quality, as illustrated in Fig.~\ref{fig:syn_data_augmentation}.
Through the two-component GMM, we can infer the posterior probability of one synthesized face being of high quality, and obtain the photorealistic faces by $p(c=1|s) > p(c=0|s)$ from all synthesized faces.

After selecting the high-quality synthesized faces, we then fine-tune the last linear layer for AIFR while keeping the convolutional layers fixed.
In long-tailed learning, rebalancing data distribution may lead to unstable training. We highlight that \seldaname would not change the data distribution too much as there are only 10\% synthesized faces in new training data. If this problem indeed happens, one can turn to the distribution alignment technique~\cite{zhang2021distribution} in long-tailed learning to stabilize the training.
We also note that this step would not influence FAS since the last linear layer was not involved in the synthesizing process of our multi-task learning framework, which will be detailed in next section.

\subsection{Multi-task Learning Framework}

This section presents our \methodname including AIFR and FAS.

\subsubsection{Age-invariant face recognition (AIFR) task}

To encourage AFD to robustly decompose features, we use an age estimation task and a face recognition task to supervise the feature decomposition. Specifically, $\mat{X}_{\mathrm{age}}$ draws the age variations through an age estimation task while $\mat{X}_{\mathrm{id}}$ encodes the identity-related information. First, as illustrated in Fig.~\ref{fig:framework}(d) we include an age estimation network $A$ with two linear layers of 512 and 101 neurons to achieve age regression similar to deep expectation~(DEX)~\cite{rothe2015dex,Rothe-IJCV-2018}, which learns the age distribution by computing a softmax expected value. Second, we append another linear layer $\mat{W} \in \mathbb{R}^{101 \times n_g}$ on top of $A$ for age classification, regularizing the learned distribution, where $n_g$ denotes the number of age groups. The loss function to optimize age estimation can be defined as:
\begin{align}
    \ell_{\textsc{ae}}(\mat{X}_{\mathrm{age}}) =\mathbb{E}_{\mat{I}}\big[&\ell_{\textsc{mse}}\left(\mathrm{DEX}({A}(\mat{X}_{\mathrm{age}})), y_\mathrm{age}\right) \notag\\
    &+ \ell_{\textsc{ce}}\left({A}(\mat{X}_{\mathrm{age}})\mat{W}, c_{\mathrm{age}}\right)\big],
\end{align}
where $y_\mathrm{age}$, $c_{\mathrm{age}}$, $\ell_{\textsc{mse}}$, and $\ell_{\textsc{ce}}$ are the ground truth age,  ground truth age group,  mean squared error (MSE) for age regression, and cross-entropy (CE) loss for age group classification, respectively. 

Next, we leverage one linear layer $L$ of 512 neurons to extract the feature vectors, and use the CosFace loss to supervise the learning of $\mat{X}_{\mathrm{id}}$ for identity classification. We also introduce a cross-age domain adversarial learning that encourages $\mat{X}_{\mathrm{id}}$ to be age-invariant through a continuous domain adaptation~\cite{wang2020continuously} with a gradient reversal layer~(GRL)~\cite{ganin2016domain}. Fig.~\ref{fig:framework}(e) shows these two components.
The final loss for AIFR is formulated as: 
\begin{align}
\mathcal{L}^{\textsc{aifr}} =& 
\ell_{\textsc{cosface}}(L(\mat{X}_{\mathrm{id}}), y_\mathrm{id})  \notag\\
+&\lambda^{\textsc{aifr}}_{\mathrm{age}}\mathcal{L}_{\textsc{ae}}(\mat{X}_{\mathrm{age}})+
\lambda_{\mathrm{id}}^{\textsc{aifr}} \mathcal{L}_{\textsc{ae}}(\mathrm{GRL}(\mat{X}_{\mathrm{id}})),
\end{align}
where the first term is the CosFace loss; the second term is the age estimation loss; the last term is the domain adaptation loss; $y_\mathrm{id}$ is the identity label; and $\lambda_{\mathrm{*}}$ controls the balance of different loss terms. Note that the second and third terms use the same network structure but have different inputs and are trained independently. The activation functions and batch normalizations are ignored for simplicity, and our face recognition model is designed strictly following the setting in~\cite{deng2019arcface} except the AFD.

\subsubsection{Face age synthesis~(FAS) task}

Fig.~\ref{fig:framework}(f) and~(g) demonstrate the FAS process of our proposed method.
In detail, the decoder $D$ adopts the architecture of the StyleGAN-based generator~\cite{karras2019style,karras2020analyzing}, which receives the discriminative facial representations $\mat{X}_{\mathrm{id}}$ at a coarse level. Then, the single-level identity-level age condition is derived from the discriminative facial representations $\mat{X}_{\mathrm{id}}$ and the same level high-resolution features extracted from the encoder $E$ using two ResBlocks~\cite{he2016deep}. Similar to the feature pyramid network~\cite{lin2017feature}, the multi-level identity-level age conditions are formed by stacking multiple ICMs at different levels. Finally, the decoder $D$ reconstructs the progressed/regressed faces from $\mat{X}_{\mathrm{id}}$ under the control of the multi-level learned identity-level age conditions, using two StyleGAN block~\cite{karras2019style,karras2020analyzing} at each level with AdaIN normalization layers~\cite{huang2017arbitrary}. Formally, the process of rendering the input face $\mat{I}$ to the synthesized face $\mat{\widehat{I}}_t$ that belongs to the target age group $t$ can be written as:
\begin{align}
    \mat{C}_t^1 &= f_1(\mat{X}_{\mathrm{id}}, \mat{E}^1, t), \\
    \mat{C}_t^l &= f_l(\mat{C}_t^l, \mat{E}^l, t), \quad l\in \{2,3\} \\
    \mat{\widehat{I}}_t &= D(\mat{X}_{\mathrm{id}}, \{\mat{C}_t^l\}_{l=1}^{3}),
\end{align}
where $l$ denotes the index of different levels, and $\mat{C}_l$ and $f_l$ are the identity-level age condition and ICMs at the $l$-th level, respectively. We note that $D$ employs several convolutional layers at each level to map the age conditions into the style latent space of StyleGAN. The rationale is that the coarse-level of $\mat{X}_{\mathrm{id}}$ corresponds to the shape of faces, \ie identities, while the fine-level of age conditions renders the faces with detail aging/rejuvenation effects.

To facilitate the visual quality of the generated faces, the FAS task is trained using GANs framework. 
In this paper, we adopt a combination of PatchDiscriminator~\cite{isola2017image} and a StyleGAN-based discriminator~\cite{karras2019style,karras2020analyzing} as our discriminator $D_\mathrm{img}$ to emphasize the local-patches of generated and real images. Furthermore, the least-squares GANs~\cite{mao2017least} are employed to optimize the GANs framework for improved quality of the generated images and stable training process, which can be formulated as follows:
\begin{align}
    \mathcal{L}_{\mathrm{adv}}^{\textsc{fas}}=\frac{1}{2} \mathbb{E}_{\mat{I}}\left[D_\mathrm{img}(\mat{\widehat{I}}_t, c^t)-1\right]^{2},
\end{align}
where $c^t$ is a scalar used in the traditional cGANs framework to align with the age condition by specifying the logits of $t$-th target age group.
To preserve the identities of input faces and improve the age accuracy, we leverage the encoder $E$, AFD, and the linear layer $L$ to supervise the FAS task. Consequently, we can achieve both face aging and rejuvenation in a holistic, end-to-end manner, as illustrated in Fig.~\ref{fig:framework}. This process can be formulated as follows:
\begin{align}
    &\mat{X}_{\mathrm{age}}^t, \mat{X}_{\mathrm{id}}^t = \mathrm{AFD}\big(E(\mat{\widehat{I}}_t)\big), \\
    &\mathcal{L}_{\mathrm{age}}^{\textsc{fas}} 
    = \ell_{\textsc{ce}}\big({A}(\mat{X}_{\mathrm{age}}^t)\mat{W}, c_{\mathrm{age}}^t\big), \\
    &\mathcal{L}_{\mathrm{id}}^{\textsc{fas}} = \mathbb{E}_{\mat{X}_{s}}\left\|\mat{X}_{\mathrm{id}}^t-\mat{X}_{\mathrm{id}}\right\|^2_F - \mathrm{cos}(L(\mat{X}_{\mathrm{id}}^t), L(\mat{X}_{\mathrm{id}})),
\end{align}
where $\|\cdot\|_F$ represents the Frobenius norm, and $\mathrm{cos}(\cdot)$ denotes cosine similarity. 
Another learned perceptual image patch similarity (LPIPS) loss~\cite{zhang2018unreasonable} is employed to further maintain the perceptual consistency between input and synthesized faces:
\begin{align}
    \mathcal{L}_{\mathrm{lpips}}^{\textsc{fas}} &= \mathbb{E}_{\mat{I}} \left[ \mathrm{LPIPS}(\mat{I}_t, \mat{\widehat{I}}_t)\right],
\end{align}

The final loss to optimize this task can be written as:
\begin{align}
    \mathcal{L}^{\textsc{fas}} = \lambda_{\mathrm{adv}}^{\textsc{fas}} \mathcal{L}_{\mathrm{adv}}^{\textsc{fas}} + \lambda_{\mathrm{id}}^{\textsc{fas}} \mathcal{L}_{\mathrm{id}}^{\textsc{fas}} + \lambda_{\mathrm{age}}^{\textsc{fas}} \mathcal{L}_{\mathrm{age}}^{\textsc{fas}} + \lambda_{\mathrm{lpips}}^{\textsc{fas}} \mathcal{L}_{\mathrm{lpips}}^{\textsc{fas}},
\end{align}
where $\lambda_{*}^{\textsc{fas}}$ controls the importance of different loss terms of FAS task. The loss function to optimize the discriminator $D_\mathrm{img}$ in the context of least-squares GANs is defined as:
\begin{align}
    \mathcal{L}_{D_{\mathrm{img}}}^{\textsc{fas}} = \frac{1}{2} \mathbb{E}_{\mat{I}_t} \left[D_{\mathrm{img}}\big(\mat{I}_t,c^t\big)-1\right]^{2} 
    +
    \frac{1}{2} \mathbb{E}_{\mat{I}} \left[D_{\mathrm{img}}\big(\mat{\widehat{I}}_{t},c^t\big)\right]^{2}.
\end{align}

At the testing stage, the only difference from existing FAS methods is that our method needs to specify the corresponding group of filters. 
Consequently, our method enjoys the advantages similar to~\cite{he2019s2gan} that the computational cost can be significantly reduced by encoding input faces only once, instead of the $n_g$ times needed in previous works~\cite{li2019age,liu2019attribute,wang2018face,yang2018learning,zhang2017age}, where $n_g$ is the number of age groups. In addition, \methodname can also achieve continuous face age synthesis similar to~\cite{yang2019learning} thanks to the StyleGAN-based decoder by interpolating in the style latent space.

\subsubsection{Optimization and inference}

The training algorithm of the proposed \methodname is shown in Algorithm~\ref{alg}
, which contains the following two stages.

In the first stage, the AIFR learns the discriminative facial representations and age estimation while the FAS produces the visual results that can boost the model interpretability for AIFR. Therefore, both two tasks can be jointly accomplished through a GAN-like optimization; they mutually leverage each other to boost themselves. In other words, the AIFR encourages FAS to render faces to preserve identity while FAS can facilitate the extraction of the identity-related features and boost the model interpretability for AIFR. Consequently, we alternately train these two tasks in a unified, multi-task, end-to-end framework. 
In the second stage, the well-trained model of previous stage is employed to synthesize paired faces, and those high-quality ones are selected for further fine-tuning with the proposed \seldaname as illustrated in Sec.~\ref{sec:ftsel}.

\begin{algorithm}[]\label{alg}
    \caption{
        Training Algorithm of \methodname
    }
    \textbf{Input:}\hspace{0mm} Dataset $\mathcal{D}=\{(\mat{I}, \vct{y}_{\mathrm{age}}, \vct{y}_{\mathrm{id}})\}$;  trainable functions $E$, $A$, $W$, $L$, $\{f_l\}_{l=1}^3$, $D$ and $D_\mathrm{img}$ \\
    \textbf{Output:}\hspace{0mm} The full trained model of \methodname.\\
    \Repeat{reaching max iteration}{
        Training $E$, $A$, $W$ and $L$ with $\mathcal{L}^{\textsc{aifr}}$ \\
        Training $D_\mathrm{img}$ with $\mathcal{L}_{D_{\mathrm{img}}}^{\textsc{fas}}$ \\
        Training $\{f_l\}_{l=1}^3$ and $D$ with $\mathcal{L}^{\textsc{fas}}$ \\
    }
    Fine-tuning $L$ with \seldaname \\
\end{algorithm}
\section{Experiments}\label{sec:exp}

\subsection{Data Collection}

\subsubsection{Large-scale cross-age face dataset}
\label{sec:training_data}
Current research on AIFR lacks a large-scale face dataset of millions of face images with large age gaps. To advance the development of AIFR and FAS, we collect and release a new large cross-age face dataset~(LCAF) with 1.7M faces from cross-age celebrities. The collection process for our dataset is summarized in the following three steps.
First, we use the public Azure Facial API~\cite{azure} to estimate the ages and genders of faces from the clean MS-Celeb-1M dataset provided by~\cite{deng2019arcface}. Second, we randomly select faces from a total of 5M faces to check whether the faces are correctly labeled, and manually correct them as best we can if any apparent mistakes occur; we mainly focus on young ages, under 20 that are often mislabeled by the API~\cite{azure}. Finally, a large-scale balanced age dataset is constructed by balancing both age and gender.

\begin{figure*}[t]
    \centering
    \includegraphics[width=0.75\linewidth]{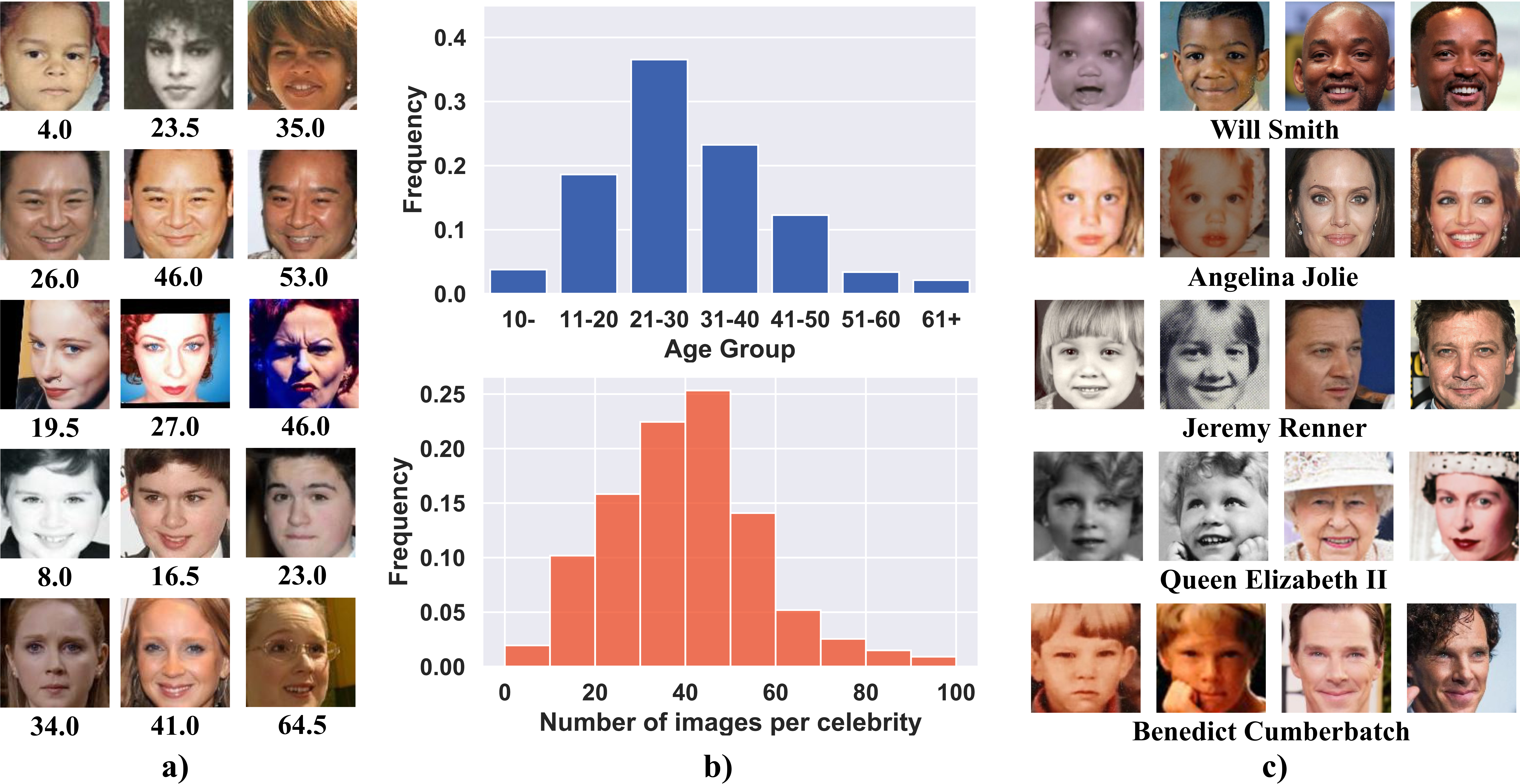}
    \caption{
        a) Sample faces and b) dataset statistics on SCAF; and c) Sample faces on \testsetname with celebrity names underneath.
    }
    \label{fig:dataset}
\end{figure*}

We further build a subset of the cross-age face dataset~(SCAF) containing approximately 0.5M images from 12k individuals following~\cite{wang2019decorrelated,wang2018orthogonal} for fair comparisons. We note that the training dataset, \ie LCAF, and the testing datasets summarized in Table~\ref{tab:testset_statistics} may have very few, or even no identities overlapping as~\cite{deng2019arcface} already removed 500+ identities from their clean MS-Celeb-1M dataset by checking the similarity of faces between the training and testing data. Fig.~\ref{fig:dataset}(a) and (b) present example images and dataset statistics of SCAF.

Following the mainstream literature~\cite{he2019s2gan,lihierarchical,li2019age,liu2019attribute,yang2018learning} with the time span of 10 years for each age group, the ages in this paper are divided into seven non-overlapping groups; \ie, 10-, 11-20, 21-30, 31-40, 41-50, 51-60, and 61+. Note that it is a much more challenging problem to perform FAS on seven groups than on the four groups widely used in previous work.  

\subsubsection{A new benchmark}

Compared to Labeled Faces in the Wild~(LFW)~\cite{huang2008labeled} dataset used in GFR, existing benchmark cross-age face datasets for AIFR such as CALFW~\cite{zheng2017cross} and AgeDB~\cite{moschoglou2017agedb}, are designed to challenge current face recognition models with cases of large age gaps. However, these datasets overlook the fact that the most challenging case of AIFR is to match faces from children to adults, which is critical for practical applications such as tracing long-missing children.

Therefore, we build a new benchmark, termed Evaluation of Cross-Age Face Recognition~(\testsetname) for AIFR, which particularly focuses on child and adult faces of the same person. Specifically, by querying the names in LFW with the keyword ``child'', we manually collect, annotate, and check the faces with the source link to ensure that the child and adult faces are from the same person. Then, we create two tasks based on \testsetname to evaluate the performance of face recognition. The first task involves 6k image pairs for $<$Adult, Child$>$, where $<$Adult, Child$>$ denotes face verification between adult and child faces. These 6k pairs are split into 10 folds, and each fold consists of 300 intra-class and 300 inter-class pairs. The second task involves 4k pairs of $<$Child, Child$>$, where the $<$Child, Child$>$ pairs are obtained with a similar procedure. Both tasks follow the protocols of LFW. We find that the second task is also challenging and worthy of further study by the face recognition community.

Table~\ref{tab:testset_statistics} summarizes the testing sets used in this paper for AIFR and Fig.~\ref{fig:dataset}(c) showcases some examples in~\testsetname. We obtain the estimated ages in the same way as stated in~Sec.~\ref{sec:training_data}. \testsetname has much larger age gaps than LFW~\cite{huang2008labeled}, CACD-VS~\cite{chen2015face}, CALFW~\cite{zheng2017cross}, and AgeDB~\cite{moschoglou2017agedb}, and has more identities and faces than FG-NET~\cite{fgnet}. In summary, there are three important properties of \testsetname: 1) clean labels, as we manually check all faces and links in the dataset; 2) larger age gaps, as \testsetname focuses on face recognition between children and adults; and 3) orthogonal to existing research on face recognition, as \testsetname contains the same identities as LFW, which makes the results on \testsetname reliable since there are no overlapping identities in current widely-used training datasets such as MS-Celeb-1M.
\begin{table}[]
    \centering
    \caption{Statistics of testing sets used in this paper for AIFR task. Due to the limited number of faces, FG-NET was mainly used for face identification. Our new benchmark dataset \testsetname has larger age gaps than the other datasets from children to adults.}\label{tab:testset_statistics}
    \begin{tabular}{lrrrrr}
    \toprule
    Dataset  & Subjects & Images & Pairs & \begin{tabular}[c]{@{}l@{}}Avg Age Gap of\\ Test Set (years)\end{tabular} \\
    \midrule
    LFW~\cite{huang2008labeled}  & 5,749 & 13,233  & 6k & 11.9 \\
    CACD~\cite{chen2015face}     & 2,000 & 163,446 & 4k & 11.7 \\
    CALFW~\cite{zheng2017cross}  & 5,749  & 12,174  & 6k & 17.6 \\
    AgeDB~\cite{moschoglou2017agedb} & 568  & 16,488  & 6k & 16.8 \\
    FG-NET~\cite{fgnet}   & 82 & 1,002  & \multicolumn{2}{c}{only for identification}  \\
    \midrule
    \testsetname     & 613  &  5,265  & 6k & 41.3 \\
    \bottomrule
    \end{tabular}
\end{table}

\subsection{Implementation Details}

Following~\cite{deng2019arcface}, we adopt ResNet-50 as the encoder $E$. In the decoder $D$, the identity age condition is bilinearly upsampled and processed with multi-level high-resolution features extracted from $E$ by two ResBlocks~\cite{he2016deep}, each of which is followed by instance normalization~\cite{ulyanov2016instance} and leaky ReLU activation of 0.2 negative slope. There are four ICBs in ICM. The age conditions are derived using several stride-2 convolutional layers and one linear layer. The style dimension of the StyleGAN-based decoder is 512. For the discriminator $D_\mathrm{img}$, we build upon the StyleGAN architecture with 4 StyleBlocks, followed by spectral normalization~\cite{miyato2018spectral} and leaky ReLU activation except in the last block, yielding an $8\times 8$ confidence map. AIFR is optimized by stochastic gradient descent~(SGD) with an initial learning rate of $0.1$ and a momentum of 0.9, while the ICM, the decoder $D$, and $D_\mathrm{img}$ are trained by Adam~\cite{kingma2014adam} with a fixed learning rate of $1.0\times 10^{-4}$, $\beta_1$ of 0.9 and $\beta_2$ of 0.99 for face age synthesis. We train all models with a mini-batch size of 512 on 8 NVIDIA GTX 2080Ti GPUs, with 110k iterations for LCAF and 36k iterations for SCAF. The learning rate of AIFR is warmed up linearly from 0 to 0.1, and then reduced by a factor of 0.1, at iterations 5k, 70k, and 90k on LCAF and 1k, 20k, 23k on SCAF. The hyper-parameters in the loss functions are empirically set as follows: $\lambda^{\textsc{aifr}}_{\mathrm{age}}$ is $0.001$, $\lambda^{\textsc{aifr}}_{\mathrm{id}}$ is $0.002$, $\lambda_{\mathrm{adv}}^{\textsc{fas}}$ is $5.0$, $\mathcal{L}_{\mathrm{lpips}}^{\textsc{fas}}$ is $1.0$, $\lambda_{\mathrm{id}}^{\textsc{fas}}$ is $1.0$, and $\lambda_{\mathrm{age}}^{\textsc{fas}}$ is $0.2$. The multiplicative margin and scale factor of CosFace loss~\cite{wang2018cosface} are set to $0.35$ and $64$, respectively. All images are aligned to $112\times 112$, with five facial landmarks detected by MTCNN~\cite{zhang2016joint}, and linearly normalized to $[-1, 1]$. For \seldaname, SCAF is employed to synthesize the child faces, leading to a total of 60k high-quality synthesized faces. Afterward, the model is fine-tuned with a learning rate of $0.01$, 10k iterations, and CosFace loss.

\subsection{Evaluation on AIFR}

    
    
    
    

    

Next, we evaluate \methodname on several benchmark cross-age datasets, including CACD-VS~\cite{chen2015face}, CALFW~\cite{zheng2017cross}, AgeDB~\cite{moschoglou2017agedb}, FG-NET~\cite{fgnet}, and the proposed \testsetname, to compare with the state-of-the-art methods. Note that MORPH is excluded since the version in~\cite{wang2019decorrelated,wang2018orthogonal,zhao2019look} is prepared for commercial use only.

\subsubsection{Result on AgeDB dataset}
\begin{table}[h]
\centering
\caption{Verification rate~(\%) on AgeDB-30 dataset for AIFR.}\label{tab:agedb}
    \begin{tabular*}{0.75\linewidth}{@{\extracolsep{\fill}}lr}
        \toprule
        Method                                      & Accuracy~(\%) \\
        \midrule
        RJIVE~\cite{sagonas2017recovering}          & 55.20      \\
        VGG Face~\cite{parkhi2015deep}              & 89.89      \\
        Center Loss~\cite{wen2016discriminative}    & 93.72      \\
        SphereFace~\cite{liu2017sphereface}         & 91.70      \\
        CosFace~\cite{wang2018cosface}              & 94.56      \\
        ArcFace~\cite{deng2019arcface}              & 95.15      \\
        DAAE~\cite{lihierarchical}                  & 95.30      \\
        \midrule
        \methodname(\textbf{ours})  & \textbf{96.45} \\
        \bottomrule
    \end{tabular*}
\end{table}
AgeDB~\cite{moschoglou2017agedb} contains 16,488 face images of 568 distinct subjects with manually annotated age labels, which has four age-invariant face verification protocols under the different age gaps of face pairs: 5, 10, 20, and 30 years. Similar to the LFW~\cite{huang2008labeled}, AgeDB is split into 10 folds for each protocol, where each fold consists of 300 intra-class and 300 inter-class pairs. We strictly follow the protocol of 30 years to perform 10-fold cross-validation since the protocol of 30 years is the most challenging one. We use models trained on SCAF to evaluate the performance on AgeDB for fair comparison. Table~\ref{tab:agedb} shows the comparison results in terms of verification accuracy, demonstrating the superior performance of \methodname over state-of-the-art methods.

\subsubsection{Result on CALFW dataset}
\begin{table}[h]
\centering
\caption{Verification rate~(\%) on CALFW dataset for AIFR.}\label{tab:calfw}
    \begin{tabular*}{0.75\linewidth}{@{\extracolsep{\fill}}lr}
        \toprule
        Method        & Accuracy~(\%) \\
        \midrule
        HUMAN-Individual  & 82.32      \\
        HUMAN-Fusion      & 86.50      \\
        \midrule
        Center Loss~\cite{wen2016discriminative}   & 85.48      \\
        SphereFace~\cite{liu2017sphereface}        & 90.30      \\
        VGGFace2~\cite{cao2018vggface2}            & 90.57     \\
        ArcFace~\cite{deng2019arcface}             & 95.45     \\
        \midrule
        \methodname(\textbf{ours}) &  \textbf{95.98} \\
        \bottomrule
    \end{tabular*}    
\end{table}
Cross-age LFW~(CALFW) dataset~\cite{zheng2017cross} is designed for unconstrained face verification with large age gaps, which contains 12,176 face images of 4,025 individuals collected using the same identities in LFW. Similarly, we follow the same protocol as the LFW, where each fold consists of 600 positive and negative pairs. We train the model on LCAF to evaluate our method on this dataset, and the results are shown in Table~\ref{tab:calfw}. Particularly, our method outperforms the recent state-of-the-art AIFR methods, establishing a new state-of-the-art on CALFW.

\subsubsection{Result on CACD-VS dataset }
\begin{table}[h]
\centering
\caption{Verification rate~(\%) on CACD-VS dataset for AIFR.}\label{tab:cacdvs}
    \begin{tabular*}{0.75\linewidth}{@{\extracolsep{\fill}}lr}
        \toprule
        Method        & Accuracy~(\%) \\
        \midrule
        HFA~\cite{gong2013hidden}                 & 84.40      \\
        CARC~\cite{chen2015face}                  & 87.60      \\
        VGGFace~\cite{parkhi2015deep}             & 96.00      \\
        Center Loss~\cite{wen2016discriminative}  & 97.48     \\
        LF-CNN~\cite{wen2016latent}               & 98.50      \\
        Marginal Loss~\cite{deng2017marginal}     & 98.95      \\
        OE-CNN~\cite{wang2018orthogonal}          & 99.20      \\
        AIM~\cite{zhao2019look}                   & 99.38      \\
        DAL~\cite{wang2019decorrelated}           & 99.40      \\
        \midrule
        \methodname(\textbf{ours})    &   \textbf{99.58} \\
        \bottomrule
    \end{tabular*}
\end{table}
The cross-age celebrity dataset (CACD) contains 163,446 face images of 2,000 celebrities in the wild, with significant variations in age, illumination, pose, and so on. Since collected by search engine, CACD is noisy with mislabeled and duplicate images. Therefore, a carefully annotated version, the CACD verification subset or CACD-VS~\cite{chen2015face}, is constructed for fair comparison, which also follows the protocol of LFW. Table~\ref{tab:cacdvs} presents the comparison of the proposed method with other state-of-the-art methods on CACD-VS. Our \methodname surpasses the other state-of-the-art methods by a large margin, making an improvement of 0.18 against the second best.

\subsubsection{Result on FG-NET dataset}
\begin{table}[h]
\centering
\caption{Rank-1 identification rate~(\%) on FG-NET~(leave-one-out) dataset for AIFR. }\label{tab:fgnet}
    \begin{tabular*}{0.75\linewidth}{@{\extracolsep{\fill}}lr}
        \toprule
        Method        & Rank-1~(\%) \\
        \midrule
        Park~\etal~\cite{park2010age}                   & 37.40      \\
        Li~\etal~\cite{li2011discriminative}            & 47.50      \\
        HFA~\cite{gong2013hidden}                       & 69.00      \\
        MEFA~\cite{gong2015maximum}                     & 76.20      \\
        CAN~\cite{xu2017age}                            & 86.50      \\
        LF-CNN~\cite{wen2016latent}                     & 88.10      \\
        AIM~\cite{zhao2019look}                         & 93.20      \\
        DAL~\cite{wang2019decorrelated}                 & 94.50      \\
        \midrule
        \methodname(\textbf{ours})    & \textbf{95.00}  \\
        \bottomrule
    \end{tabular*}
\end{table}

\begin{table}[h]
\centering
\caption{Rank-1 identification rate~(\%) on FG-NET~(MF1) dataset for AIFR.}\label{tab:fgnet_mf1}
    \begin{tabular*}{0.75\linewidth}{@{\extracolsep{\fill}}lr}
        \toprule
        Method        & Rank-1~(\%) \\
        \midrule
        FUDAN-CS\_SDS~\cite{wang2017multi}          & 25.56      \\
        SphereFace~\cite{liu2017sphereface}            & 47.55      \\
        TNVP~\cite{Duong_2017_ICCV}           & 47.72      \\
        OE-CNN~\cite{wang2018orthogonal}           & 52.67      \\
        DAL~\cite{wang2019decorrelated}           & \textbf{57.92}      \\
        \midrule
        \methodname(\textbf{ours})   &  \underline{57.78} \\
        \bottomrule
    \end{tabular*}
\end{table}
FG-NET~\cite{fgnet} is the most popular and challenging age dataset for AIFR, which consists of 1,002 face images from 82 subjects collected from the wild with huge age variations ranging from child to elder. We strictly follow the evaluation pipeline in~\cite{wang2019decorrelated,wang2018orthogonal}. Specifically, the model is trained on SCAF and tested under the protocols of leave-one-out and MegaFace challenge 1~(MF1). In the leave-one-out protocol, faces are used to match the remaining faces, repeating 1,002 times. Table~\ref{tab:fgnet} reports the rank-1 recognition rate. Our method outperforms prior work by a large margin. On the other hand, the MF1 contains an additional 1M images as the distractors in the gallery set from 690k different individuals, where models are evaluated under the large and small training set protocols. The small protocol requires the training set to be less than 0.5M images; this is strictly followed to evaluate our trained model on FG-NET, and the experimental results are reported in Table~\ref{tab:fgnet_mf1}. Our method achieves competitive performance against other methods since the distractors in MF1 contain a large number of mislabeled probe and gallery face images.

\subsubsection{Result on \testsetname dataset}

\begin{table}[h]
    \centering
    \caption{Verification rate~(\%) on \testsetname dataset for AIFR. 
     }\label{tab:new_testset}
    \begin{tabular*}{1\linewidth}{@{\extracolsep{\fill}}lcc}
        \toprule
        Method        & $<$Adult, Child$>$ & $<$Child, Child$>$ \\
        \midrule
        Human, Average & 73.34 & 68.62 \\
        Human, Voting  & 85.95 & 78.75 \\
        \midrule
        Softmax        & 85.03 & 88.25 \\
        CosFace~\cite{wang2018cosface}        & 85.72 & 90.75 \\
        ArcFace~\cite{deng2019arcface}        & 86.52 & 90.65 \\
        CurricularFace~\cite{huang2020curricularface}        & 84.78 & 90.80 \\
        \midrule
        \methodname(\textbf{ours})  & \textbf{87.55} & \textbf{91.20} \\
        \bottomrule
    \end{tabular*}
\end{table}

\testsetname is the proposed cross-age benchmark for AIFR, which contains 5,265 faces including paired child and adult faces from 613 subjects. It follows the protocols of LFW and contains subset identities of LFW, which aims to evaluate the performance of face verification based on two tasks, \ie $<$Adult, Child$>$ and $<$Child, Child$>$ with 6k and 4k pairs, respectively.

To establish baseline results on \testsetname, we first evaluate humans on the task of distinguishing two faces of child and adult from the same person, following the procedures of LFW~\cite{huang2008labeled} and CACD-VS~\cite{chen2015face}. Specifically, we ask the users at Amazon Mechanical Turk~(AMT)~\cite{amt} to determine whether a given pair of faces belongs to the same person, and provide their confidence in the choices. To obtain more confident results, the users must have at least 95\% approval rate at AMT, and each pair of faces has 10 results from different users. We report the human performance in Table~\ref{tab:new_testset}. The voting performance significantly surpasses the average performance as the voting procedure ensembles the results from multiple users.

\begin{figure}[h]
    \centering
    \includegraphics[width=1\linewidth]{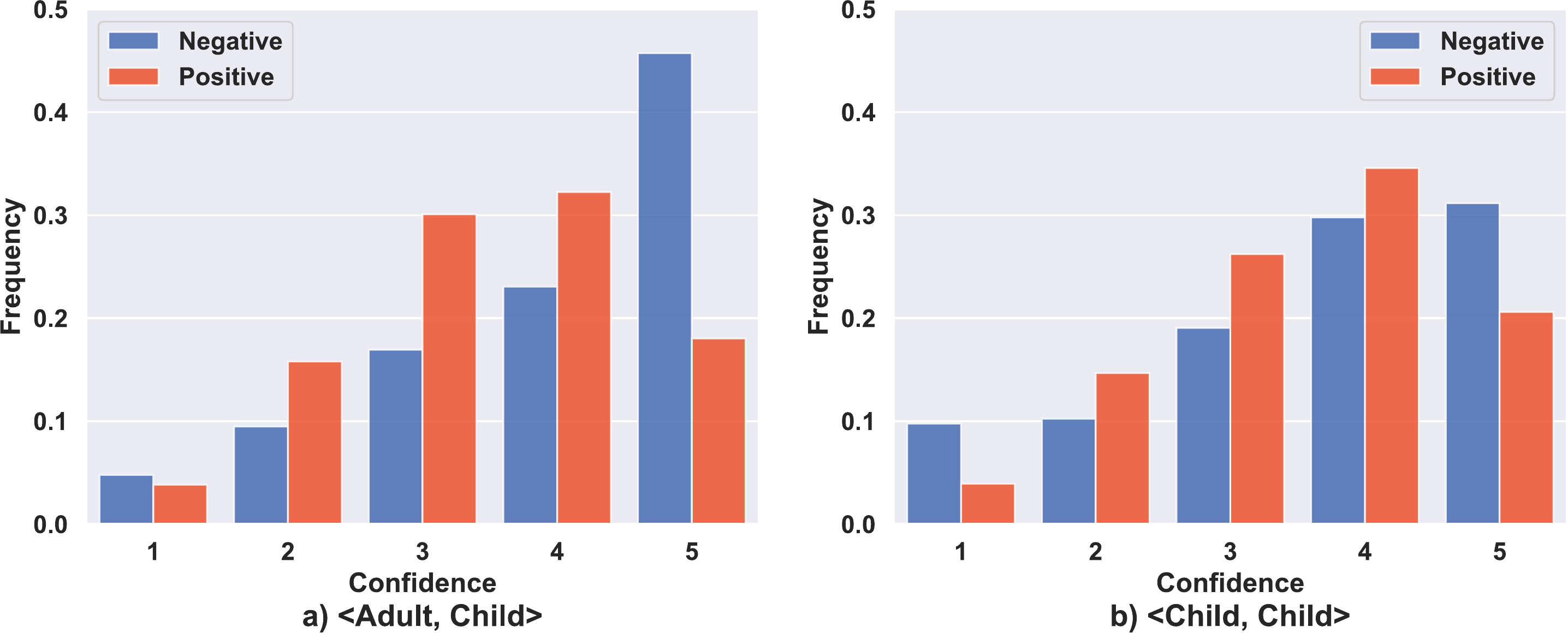}
    \caption{
        Visualization of the confidence scores of humans regarding their choices, where a higher value indicates greater confidence.
        }
    \label{fig:human_testset}
\end{figure}


We then reproduced several baseline face recognition methods including Softmax, CosFace~\cite{wang2018cosface}, ArcFace~\cite{deng2019arcface}, and CurricularFace~\cite{huang2020curricularface}, and report the results of our \methodname. Not surprisingly, \methodname outperforms humans and other baseline face recognition methods. It is interesting to observe that the performance of humans on $<$Adult, Child$>$ and $<$Child, Child$>$ has the opposite results to those of CNN-based methods, \ie, the results of $<$Adult, Child$>$ are better than the ones of $<$Child, Child$>$. A possible reason is that humans are not so confident in the choices of negative pairs in $<$Child, Child$>$ compared to $<$Adult, Child$>$, as shown in Fig.~\ref{fig:human_testset}. To obtain more insights into the AIFR task on \testsetname, in supplemental Fig.~\ref{fig:case_analysis}, we showcase some example pairs with their cosine similarities on \testsetname for two sub-tasks.


\subsubsection{Ablation study}

\begin{table}[h]
\centering
\caption{Ablation study on the components of \methodname for AIFR.}\label{tab:ablation_study}
    \scalebox{0.92}{
    \begin{tabular}{lccccc}
        \toprule
        Model & AgeDB-30 & CALFW & CACD-VS & FG-NET \\ 
        \midrule
        Baseline & 95.52 & 94.27 & 99.12 & 93.64 \\
        \quad+Age & 95.32 & 94.35 & 99.15 & 93.88 \\ 
        \quad+AFD~(CA) & 95.63 & 94.50 & 99.32 & 94.05 \\ 
        \quad+AFD~(SA) & 95.85 & 94.43 & 99.25 & 94.38 \\ 
        \quad+AFD~(CBAM) & 96.08 & 94.32 & 99.18 & 94.36 \\ 
        \quad+AFD & 95.90 & 94.48 & 99.30 & 94.58 \\ 
        \methodname w/o FT & \underline{96.23} & \underline{94.72} & \underline{99.38} & \underline{94.78}\\ 
        \methodname w/~~ FT-All & 95.88 & 94.45 & 99.22 & 93.98 \\ 
        \midrule
        \methodname~w/~~ \seldaname   & \textbf{96.45} & \textbf{94.97} & \textbf{99.45} & \textbf{95.00} \\
        \bottomrule
    \end{tabular}
    }
\end{table}

To investigate the efficacy of different modules in~\methodname, we perform ablation studies based on four benchmark datasets for AIFR by considering the following variants of our method.
1) Baseline: we remove all extra components other than the CosFace loss to train the face recognition model. 
2) +Age: this variant is jointly trained under the supervision of both CosFace and age estimation loss, similar to~\cite{wang2019decorrelated,zheng2017age}. 
3) +AFD~(CA), +AFD~(SA), +AFD~(CBAM), +AFD: these four variants utilize the proposed attention-based feature decomposition to highlight the age-related information at different levels with different attention modules: CA~\cite{hu2018squeeze}, SA~\cite{woo2018cbam}, CBAM~\cite{woo2018cbam}, and the proposed one. 
4) \methodname w/o FT: our proposed~\methodname is trained simultaneously by the AFD and cross-age domain adaptation loss without fine-tuning on the synthesized faces, which is the same as our preliminary version in~\cite{huang2021age}.
5) \methodname w/ FT-All: directly using all the synthesized faces for fine-tuning.
6) \methodname w/ \seldaname: using \seldaname to select high-quality synthesized faces for fine-tuning.

Table~\ref{tab:ablation_study} presents the experimental results. 
Note that the verification rate of the baseline model on AgeDB-30 is higher than those of ArcFace and DAAE since our training data are age-balanced, which is an important feature of our collected dataset. Even though the age estimation task is performed in the face recognition model, it cannot introduce any improvement in AIFR compared to the baseline model. On the other hand, AFD achieves a remarkable performance improvement on all cross-age datasets. Nevertheless, as AFD highlights the age-related information at both the channel and spatial levels in parallel, our method achieves consistent performance improvements, demonstrating its effectiveness compared to the single level~(CA and SA) or sequential level~(CBAM). Furthermore, the use of cross-age domain adversarial training leads to an additional performance improvement.
At last, when all synthesized child faces are employed as the additional training data~(\ie~\methodname w/ FT-All), the performance drops as 1) the artifacts in low-quality synthesized faces would degrade the AIFR model, which has also been reported in~\cite{wang2018face}; and 2) using all synthesized children' faces make the face recognition model focus more on children, where there are most adult faces in the testing sets.
However, applying \seldaname to AIFR~(\ie~\methodname w/ \seldaname) has consistently improved the performance on several benchmark datasets.
In supplemental Table~\ref{tab:comparisons_to_conf}, we show that \seldaname has significantly improved the performance on children under 10 years old~$<$Child, Child$>$, and across ages~$<$Adult, Child$>$ on the ECAF over our preliminary version~\cite{huang2021age} without synthesized faces.

\subsection{Evaluation on GFR}

\begin{table}[h]
\centering
\caption{General face recognition on LFW and MF1-Facescrub dataset.}\label{tab:gfr}
    \begin{tabular*}{0.75\linewidth}{@{\extracolsep{\fill}}lcc}
        \toprule
        Method        & LFW   & MF1-Facescrub \\
        \midrule
        SphereFace~\cite{liu2017sphereface}    & 99.42 & 72.73         \\
        CosFace~\cite{wang2018cosface}         & 99.33 & 77.11         \\
        OE-CNN~\cite{wang2018orthogonal}       & 99.35 & N/A           \\
        DAL~\cite{wang2019decorrelated}        & \underline{99.47} & \textbf{77.58}         \\
        \midrule
        \methodname(\textbf{ours})  & \textbf{99.55} & \underline{77.33}  \\
        \bottomrule
    \end{tabular*}
\end{table}

To validate the generalization ability of our \methodname for GFR, we further conduct experiments on the LFW~\cite{huang2008labeled} and MegaFace Challenge 1 Facescrub~(MF1-Facescrub)~\cite{kemelmacher2016megaface} datasets. LFW~\cite{huang2008labeled} is the most popular public benchmark dataset for GFR, which contains 13,233 face images from 5,749 subjects. MF1-Facescrub~\cite{kemelmacher2016megaface} uses the Facescrub dataset~\cite{ng2014data} of 106,863 face images from 530 celebrities as a probe set. The most challenging problem of MF1 is that it uses an additional 1M face images in the gallery set as distractors in face matching. That is, the results on MF1 are not as reliable as those on LFW due to the extremely noisy distractors in MF1. We strictly follow the same procedure as~\cite{wang2019decorrelated,wang2018orthogonal}; \ie, the training dataset contains 0.5M images~(SCAF). 

Table~\ref{tab:gfr} reports the verification rate on LFW and the rank-1 identification rate on MF1-Facescrub against the state-of-the-art GFR methods. Our method achieves competitive performance on both datasets, demonstrating the strong generalization ability of our~\methodname. We highlight that our \methodname can provide photorealistic synthesized faces to improve model interpretability, which is absent in other methods~\cite{wang2019decorrelated,wang2018orthogonal}.

\subsection{Evaluation on FAS}\label{sec:qualitative_comparison}
We further evaluate the model trained on SCAF for FAS.

\subsubsection{Qualitative results}

\begin{figure*}[t]
    \centering
    \includegraphics[width=0.9\linewidth]{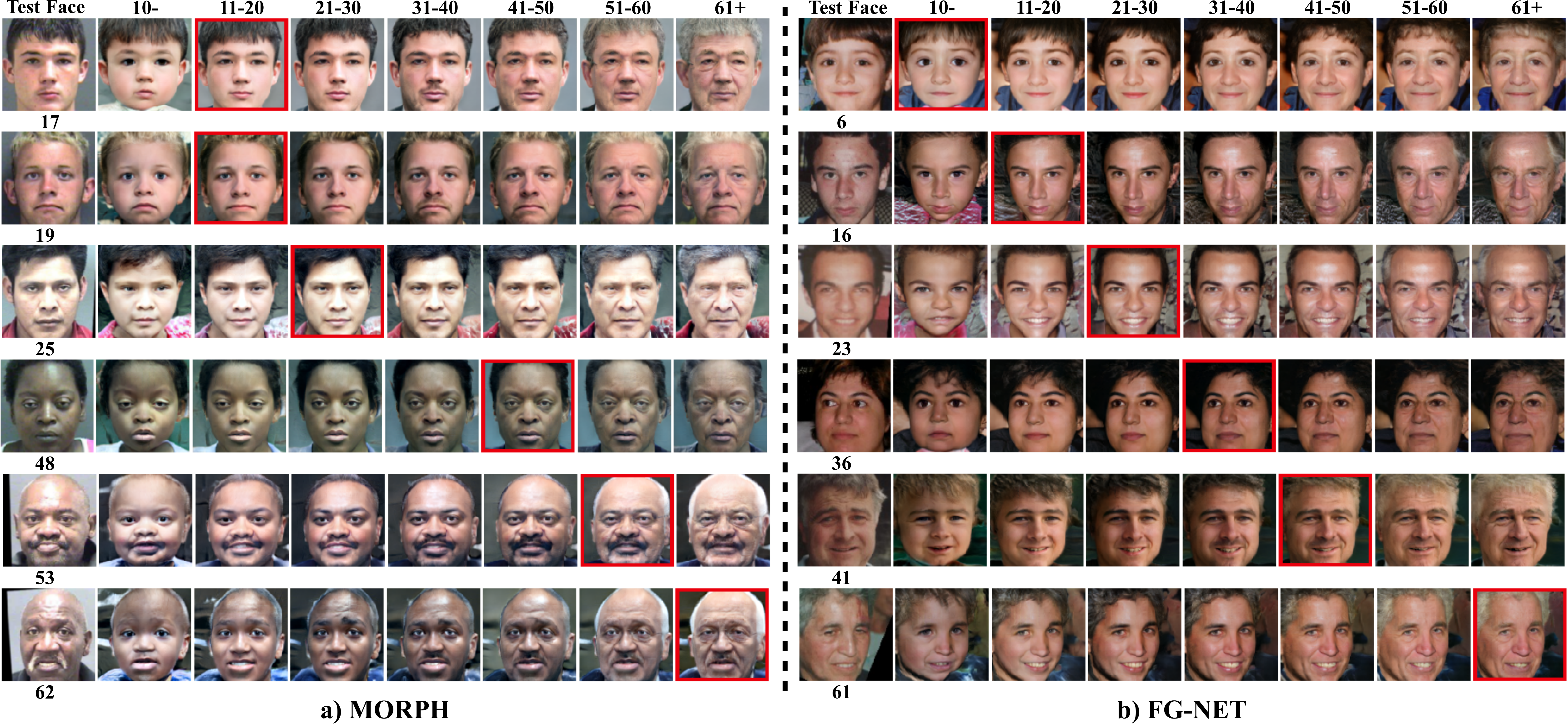}
    \caption{Qualitative results by applying our \methodname trained on the SCAF dataset to two external datasets : a) MORPH; and b) FG-NET. Red boxes indicate input faces.}
    \label{fig:qualitative_results}
\end{figure*}

Fig.~\ref{fig:qualitative_results} presents some sample results on the external datasets including MORPH and FG-NET. Our method is able to simulate the face age synthesis process between age groups with high visual fidelity. Although variations exist in terms of race, gender, expression, and occlusion, the synthesized faces are still photorealistic, with natural details in the skin, muscles, and wrinkles while consistently preserving identities, thus confirming the generalization ability of the proposed method.

\subsubsection{Comparisons with prior work}

\begin{figure*}[t]
    \centering
    \includegraphics[width=0.95\linewidth]{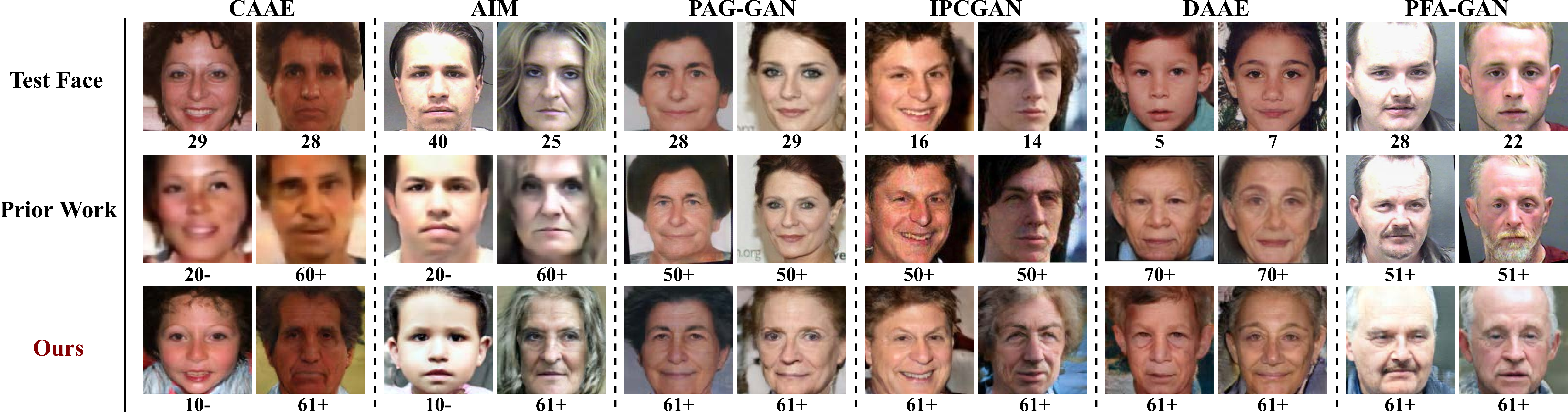}
    \caption{Comparisons with prior work on FG-NET, MORPH, and CACD datasets. We show the test faces in the first row with the real age labels below each image. The second row presents two sample results of prior work, with the target age below each image. The third row shows our results of the same input faces for the same face aging and rejuvenation as in prior work. Zoom in for a better view of image details.}
    \label{fig:qualitative_comparison}
\end{figure*}

We also conduct qualitative comparisons with prior work including CAAE~\cite{zhang2017age}, AIM~\cite{zhao2019look}, PAG-GAN~\cite{yang2018learning}, IPCGAN~\cite{wang2018face}, DAAE~\cite{lihierarchical}, and PFA-GAN~\cite{huang2020pfa} on MORPH, FG-NET, and CACD datasets. Fig.~\ref{fig:qualitative_comparison} shows that both CAAE and AIM produce oversmoothed faces due to their image reconstruction and other superior methods age faces with more photorealistic aging effects but subtle changes. However, our \methodname can synthesize faces with facial shape changes and stronger aging effects even though the age gaps are large~(see the comparisons with DAAE). In addition to the image quality, \methodname has not been re-trained on these three benchmark datasets and owns finer age split~(7 versus 4 age groups in most previous literature), which demonstrates the strong generalization ability of \methodname against other competitors. Note that the results of competitors are directly referred from their own papers for a fair comparison, which is widely adopted in the FAS literature such as~\cite{he2019s2gan,lihierarchical,li2019age,liu2019attribute,yang2018learning} to avoid any bias or error caused by self-implementation. 

\subsubsection{Quantitative comparisons}

\begin{table*}[t]
    \centering
    \caption{Quantitative comparisons between our MTLFace and the state-of-the-art face aging/rejuvenation methods in the form of $a/b/c$, where $a$, $b$, and $c$ represent the mean values of age accuracy~(\%), mean absolute error, and identity preservation (cosine similarity) computed over all age mappings, respectively.
    }
    \begin{tabular*}{0.90\linewidth}{@{\extracolsep{\fill}}lcccc}
    \toprule
     Method       & MORPH       & FG-NET      & CACD & \testsetname       \\ 
    \midrule
    CAAE~\cite{zhang2017age}    & 39.77/8.83/0.131\std{0.089} & 40.72/9.32/0.146\std{0.097} & 39.50/8.74/0.118\std{0.092} & 41.36/8.75/0.139\std{0.100} \\
    IPCGAN~\cite{wang2018face}  & 58.44/4.95/0.608\std{0.109} & 61.22/4.32/0.453\std{0.122} & 62.04/3.99/0.452\std{0.129} & 61.92/4.16/0.511\std{0.133} \\ 
    S2GAN~\cite{he2019s2gan}    & 53.85/5.47/0.369\std{0.105} & 60.26/4.37/0.293\std{0.111} & 60.19/4.31/0.270\std{0.110} & 62.35/3.98/0.297\std{0.117} \\ 
    \midrule
    MTLFace~(\textbf{ours}) & \textbf{67.07}/\textbf{2.98}/\textbf{0.652}\std{0.091} & \textbf{71.33}/\textbf{3.10}/\textbf{0.648}\std{0.094} & \textbf{71.94}/\textbf{2.55}/\textbf{0.630}\std{0.096} & \textbf{72.14}/\textbf{2.28}/\textbf{0.620}\std{0.096} \\

    \quad w/o AIFR & 51.79/5.49/0.219\std{0.095} & 48.60/7.68/0.235\std{0.093} & 50.69/6.50/0.197\std{0.095} & 52.31/6.14/0.217\std{0.103} \\ 

    \quad w/o Multi-level ICMs & 64.80/3.35/0.606\std{0.090} & 69.35/3.27/0.628\std{0.094} & 67.79/3.02/0.613\std{0.098} & 70.77/2.65/0.601\std{0.099} \\ 

    \quad w/o ICM & 62.60/3.86/0.605\std{0.072} & 65.46/3.65/0.612\std{0.080} & 62.55/3.87/0.591\std{0.099} & 67.34/3.16/0.584\std{0.088} \\
    \bottomrule
    \end{tabular*}
    \label{tab:fas_quantitative_results}
\end{table*}

We quantitatively evaluate \methodname over previous competitive methods including CAAE~\cite{zhang2017age}, IPCGAN~\cite{wang2018face}, and S2GAN~\cite{he2019s2gan}, by the following three metrics:
\begin{enumerate}
\item[1)] age accuracy: we trained a ResNet-100 model on 80\% faces of LCAF using $\ell_{\textsc{ae}}$ as the loss function to predict the ages of all synthesized faces, where the proportion of the predicted ages falling into the target age groups is the age accuracy;
\item[2)] mean absolute error~(MAE): MAE between predicted and ground-truth ages used in~\cite{tan2017efficient,zhu2021improving,Rothe-IJCV-2018} is adopted as one of evaluation metric. Here, the mean ages of target age groups is the ground-truth age label for synthesized faces; \eg, 5 years old for 10- age group, similar to~\cite{zhu2020look}; and
\item[3)] identity preservation: an external well-trained face recognition model, the ResNet-100 network pre-trained on the MS-Celeb-1M dataset from~\cite{deng2019arcface}, is used for fair comparisons to compute the cosine similarity between the input and synthesized faces.
\end{enumerate}

We reproduce CAAE, IPCGAN, and S2GAN on the SCAF dataset for fair comparisons, and then directly apply them to three external cross-age datasets: MORPH\cite{ricanek2006morph}, FG-NET~\cite{fgnet}, CACD~\cite{chen2015face}, and the proposed \testsetname. Table~\ref{tab:fas_quantitative_results} presents the quantitative results of different face aging/rejuvenation methods, including the competitive methods, our proposed MTLFace and its two variants~(w/o multi-level ICMs and w/o ICM), in terms of three evaluation metrics. MTLFace outperforms other competitors by a clear margin; this is a direct result of AIFR and ICM with multi-level architecture. Without ICM, MTLFace reduces to a common cGANs-based method that uses one-hot encoding to control face aging/rejuvenation at the group level. Remarkably, the MTLFace without ICM still outperforms these two baseline methods, implying that our multi-learning framework with attention-based feature decomposition is effective in improving the quantitative results. 
We also ablate AIFR from \methodname for FAS task; \ie, the encoder was trained with FAS without the training signals of AIFR. Without AIFR, FAS cannot produce faces with the same identity at the target age group as expected, which shows the essential importance of the AIFR task for FAS;
see supplemental Fig.~\ref{fig:qualitative_comparison_wo_AIFR} for qualitative comparisons between \methodname with and without AIFR.

\subsubsection{Comparison with ground-truth faces}
\label{sec:gt_comparison}
\testsetname provides pairs of child and adult faces from the same person, which enables a comparison between the synthesized faces and ground-truth faces.
To this end, we also conduct comparisons with the trained CAAE, IPCGAN, S2GAN, and one of the current state-of-the-art methods LATS~\cite{or2020lifespan}. Note that we use the public pre-trained model of LATS to produce the synthesized faces,  which are then aligned with MTCNN~\cite{zhang2016joint} to avoid potential self-implementation bias.

\begin{figure}[h]
    \centering
    \includegraphics[width=1.0\linewidth]{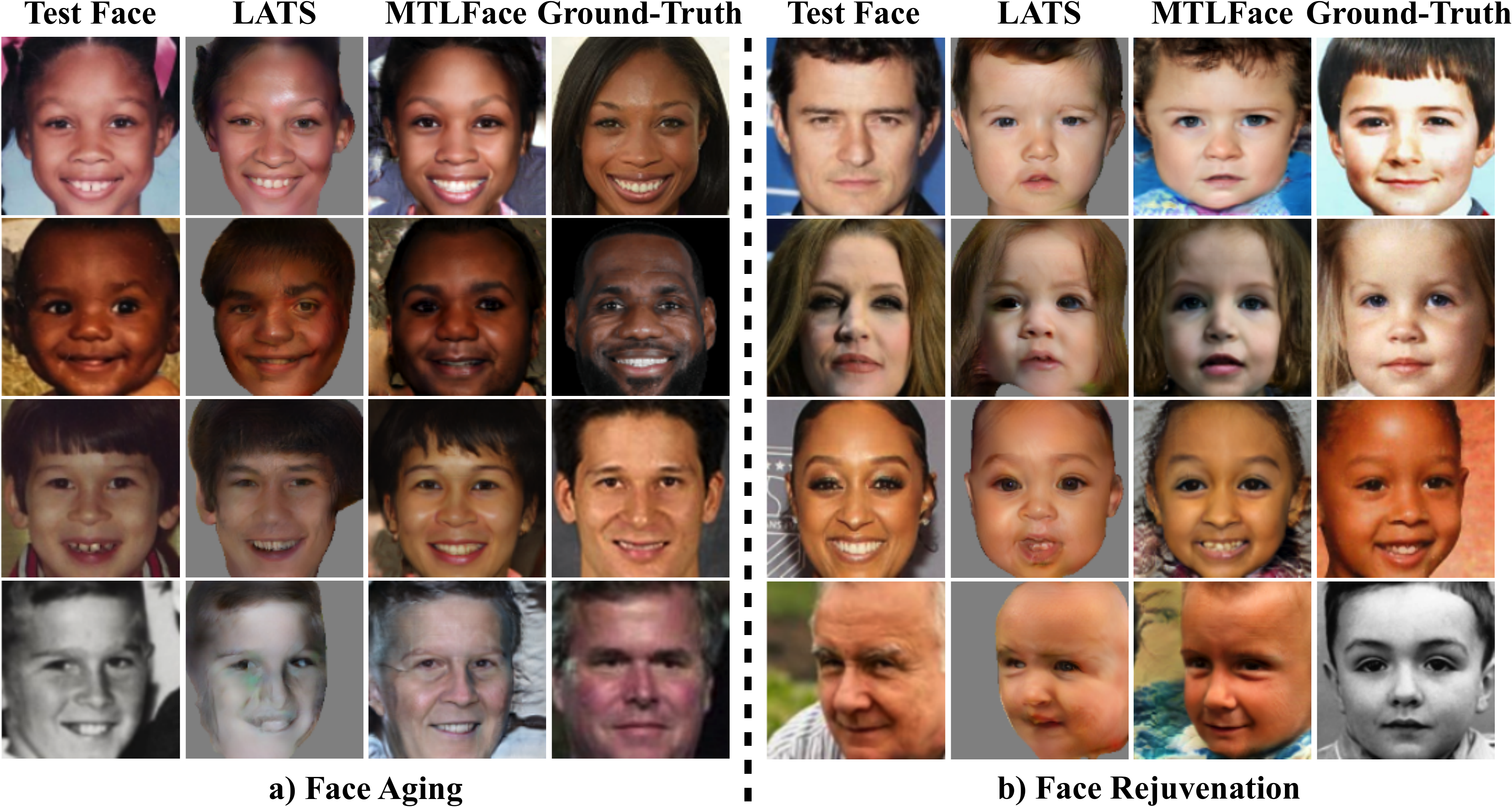}
    \caption{Comparison with ground-truth images on \testsetname. We synthesize the faces by a) face aging and b) face rejuvenation to the same age group.}
    \label{fig:gt_aging_rejuvenation}
\end{figure}

Fig.~\ref{fig:gt_aging_rejuvenation} visualizes the aging/rejuvenation process for \methodname and LATS with the reference of ground-truth faces. Even though there are large age gaps from child to adult, \methodname can synthesize realistic faces with similar aging/rejuvenation effects compared to ground-truth faces in terms of identities. On the contrary, LATS produces faces with severe artifacts for these low-quality input faces---low resolution and extreme poses---which indicates that \methodname has a stronger generalization ability.

To further demonstrate the effects of face age synthesis, we quantitatively evaluate the synthesized faces against corresponding ground-truth adult/child faces in terms of cosine similarity. First, the positive pairs from \testsetname are aged/rejuvenated to the same age groups as the ground-truth faces. Then, the trained face recognition model of CosFace is employed for fair comparisons to compute the cosine similarity between the synthesized and ground-truth faces.
The results are reported in Table~\ref{tab:comparison_gt} in terms of face aging and rejuvenation. Compared to the results of the ground-truth, the face rejuvenation of \methodname significantly improves the discrimination of the face recognition model~(see increased cosine similarity). However, it drops slightly for face aging.
Similarly, this phenomenon turns out consistently on other methods for face aging.
We think the reason is that face aging is much more difficult than face rejuvenation, so the unexpected ghost artifacts harm the recognition performance.
Although face aging is largely influenced by individuals, health conditions, personal habits, and so on, \methodname still performs better in preserving personal identities than other methods in terms of ground-truth, demonstrating the effectiveness of AIFR task.

\begin{table}[]
    \centering
    \caption{Cosine Similarity on ground-truth face pairs of \testsetname for FAS. The higher values, the better identity preservation in synthesized faces.}\label{tab:comparison_gt}
    \begin{tabular}{lcc}
    \toprule
    Method & Aging & Rejuvenation \\
    \midrule
    Ground-Truth            &  0.182\std{0.111}  & 0.182\std{0.111}\\
    \midrule
    CAAE~\cite{zhang2017age} & 0.030\std{0.077}   & 0.078\std{0.088}\\
    IPCGAN~\cite{wang2018face} &  0.137\std{0.099}  & 0.193\std{0.098}\\
    S2GAN~\cite{he2019s2gan} &  0.065\std{0.090}  & 0.135\std{0.095}\\
    LATS~\cite{or2020lifespan} &  0.103\std{0.090}  & 0.172\std{0.106} \\
    MTLFace~(\textbf{ours}) &  \textbf{0.164}\std{0.106}  & \textbf{0.210}\std{0.112}   \\
    \bottomrule
    \end{tabular}
\end{table}

\begin{figure*}[h]
    \centering
    \includegraphics[width=0.9\linewidth]{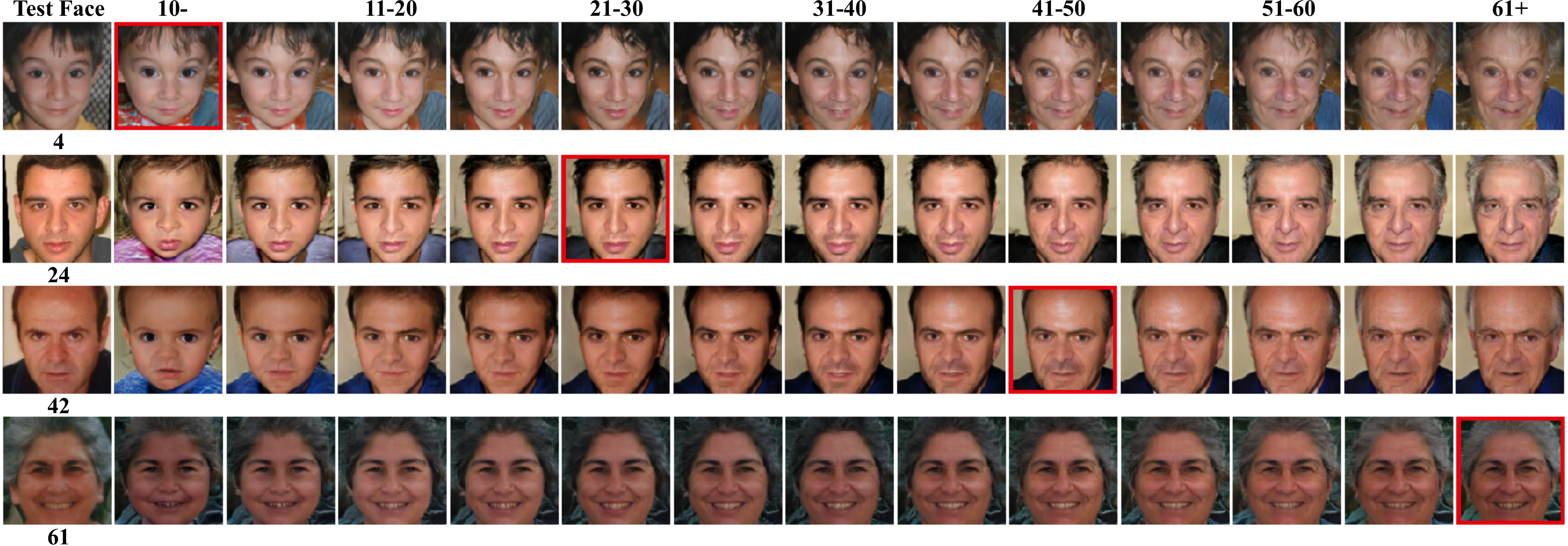}
    \caption{Sample results of continuous face age progression and regression for \methodname on the FG-NET dataset. We show the test faces in the first column, with the age below the image. The red boxes indicate the reconstructed faces in the same age groups as the input faces. We synthesize the continuous aging/rejuvenation faces by interpolating between the two latent variables of adjacent age groups. Zoom in for better view.
    }
    \label{fig:continuous_face_age_synthesis}
\end{figure*}
\subsubsection{Continuous face age synthesis}
Thanks to the StyleGAN-based architecture used in the face age synthesis task, \methodname can interpolate the missing classes in the latent space between two adjacent age groups to achieve continuous age progression and regression. Fig.~\ref{fig:continuous_face_age_synthesis} shows examples of continuous face age synthesis for \methodname. Compared to the synthesis only on discrete age groups in previous works~\cite{lihierarchical,li2019age,liu2019attribute,yang2018learning}, \methodname can directly interpolate in the latent space to synthesize realistic faces with smooth aging sequences.
Note that the continuous synthesis is \emph{jointly} achieved by the StyleGAN architecture and the proposed ICM, whereas only StyleGAN cannot perform well on FAS. In supplementary Sec.~\ref{sec:clarify_diff}, we clarify the differences between StyleGAN and our \methodname for continuous synthesis, summarize the advantages of \methodname, and conduct extensive quantitative and qualitative comparisons.

\subsubsection{Training visualization}
This section examines the training stability of \methodname. That is, when AIFR is not well trained at the early stage of training, FAS may not preserve the identities of input faces, which consequently harms the quality of synthesized faces. However, we argue that the unstable training would not arise in \methodname due to the following reasons:
\begin{enumerate}

\item[1)] In addition to the identity preservation loss, another learned perceptual image patch similarity (LPIPS) loss~\cite{zhang2018unreasonable} is employed to further encourage the perceptual consistency between input and synthesized faces. Although AIFR cannot provide FAS with a good training signal at the early stage of training, the LPIPS loss can still encourage FAS to reconstruct the input faces.

\item[2)] AIFR can extract discriminative representations that are good enough for training FAS at the beginning of training.
    
\end{enumerate}

\begin{figure}[t]
    \centering
    \includegraphics[width=1.0\linewidth]{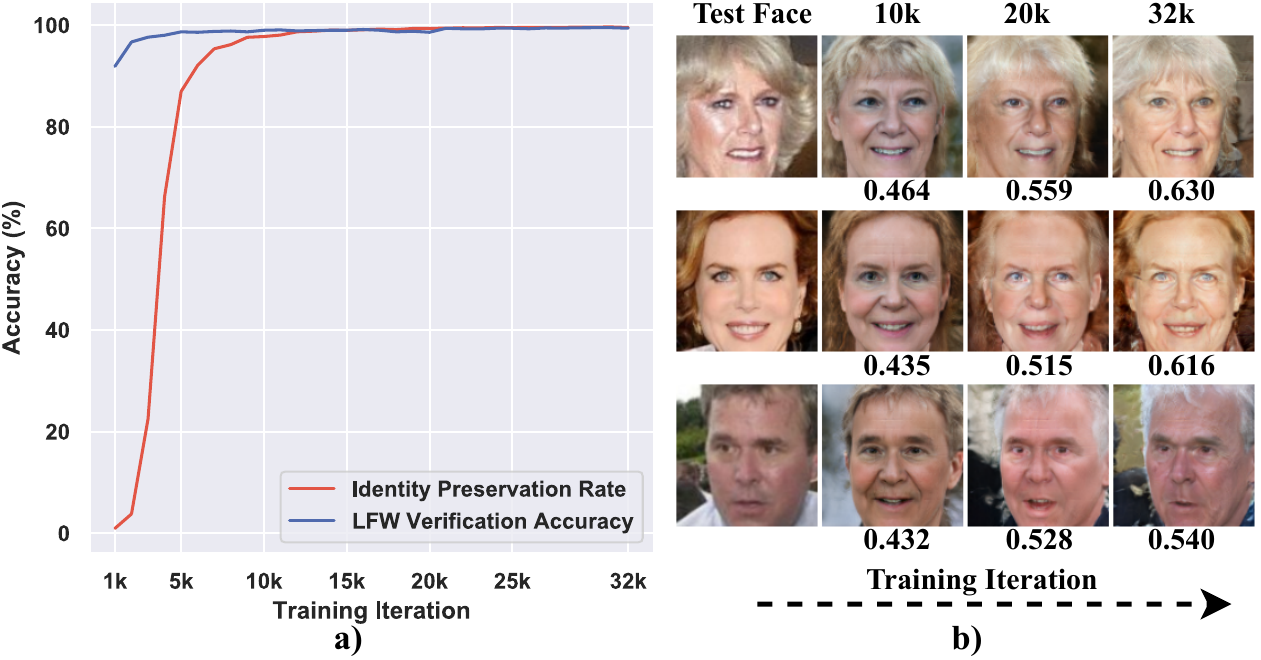}
    \caption{
        Visualization of the training process in terms of qualitative and quantitative results. a) Identity preservation rate of face age synthesis and LFW verification accuracy of face recognition across training. 
        The identity preservation rate denotes the ratio of the pairs whose cosine similarity is larger than the threshold computed on LFW.
        Even at the early stage of the training process, AIFR performs well so that it can provide good enough training signals to encourage FAS for photo-realistic faces. b) Synthesized faces of the 61+ age group at different training iterations with the cosine similarities between input and synthesized faces shown below. FAS can reconstruct pleasing faces at the early stage of training, and become better at preserving identities with the training of AIFR.
        }
    \label{fig:training_process_vis}
\end{figure}

To validate these two points, we visualize the training process of both AIFR and FAS in Fig.~\ref{fig:training_process_vis}. At the first 1k iterations, AIFR can achieve a good performance on LFW verification while FAS cannot reconstruct the same identities. The results indicate that although AIFR has not reached its best performance, it can still provide good enough training signals for FAS. In terms of synthesized faces, FAS can still reconstruct pleasing faces benefitting from the LPIPS loss, and become better at preserving the identities with the training of AIFR.

\section{Discussion}\label{sec:discussion}

Although the proposed \methodname has achieved state-of-the-art performance for the AIFR task and can produce pleasing synthesized faces for the FAS task, we acknowledge some limitations in this work. Specifically, although StyleGAN-based decoder has significantly improved the image quality of synthesized faces and achieved continuous face age synthesis, we found that it cannot well preserve the background of input faces. Since face age synthesis mainly focuses on facial changes, the background may be relatively less important. This may be addressed by the following two solutions. First solution is to use another pixel-wise loss such as mean absolute loss, which may inevitably reduce the aging/rejuvenation effects. Seond solution is to extract the face area from the input images prior to being fed into the synthesis network to reduce the influence of background like LATS~\cite{or2020lifespan}, which may, however, introduce extra computational cost for face segmentation.

\section{Conclusion}\label{sec:conc}

In this paper, we proposed a multi-task learning framework, termed \methodname, to achieve AIFR and FAS simultaneously. We proposed two novel modules: AFD to decompose the features into age- and identity-related features, and ICM to achieve identity-level FAS.
We also proposed a novel selective fine-tune strategy to boost AIFR that selects and leverages the high-quality synthesized faces.
Extensive experiments on both cross-age and general benchmark datasets for face recognition demonstrate the superiority of our \methodname. With multi-level ICMs, \methodname can be significantly better at preserving the identities of input faces by multi-level skip connections, and improve the age accuracy due to the multi-level age conditions. The newly collected large-scale cross-age training dataset and benchmark could further advance the development of AIFR and FAS.





\section*{Acknowledgments}
This work was supported in part by Shanghai Municipal Science and Technology Major Project (2018SHZDZX01), ZJ Lab,  the National Natural Science Foundation of China (62176059 and 62101136), Shanghai Municipal of Science and Technology Project (20JC1419500), and Shanghai Center for Brain Science and Brain-inspired Technology. 
The authors would like to thank the associate editor and two anonymous reviewers for their valuable comments, which greatly improved the quality of this article.

\ifCLASSOPTIONcaptionsoff
  \newpage
\fi



\clearpage
\onecolumn

\begin{center}
\huge
\textbf{
---Supplementary Material---
}
\end{center}

\begin{figure*}[h]
    \centering
    \includegraphics[width=0.8\linewidth]{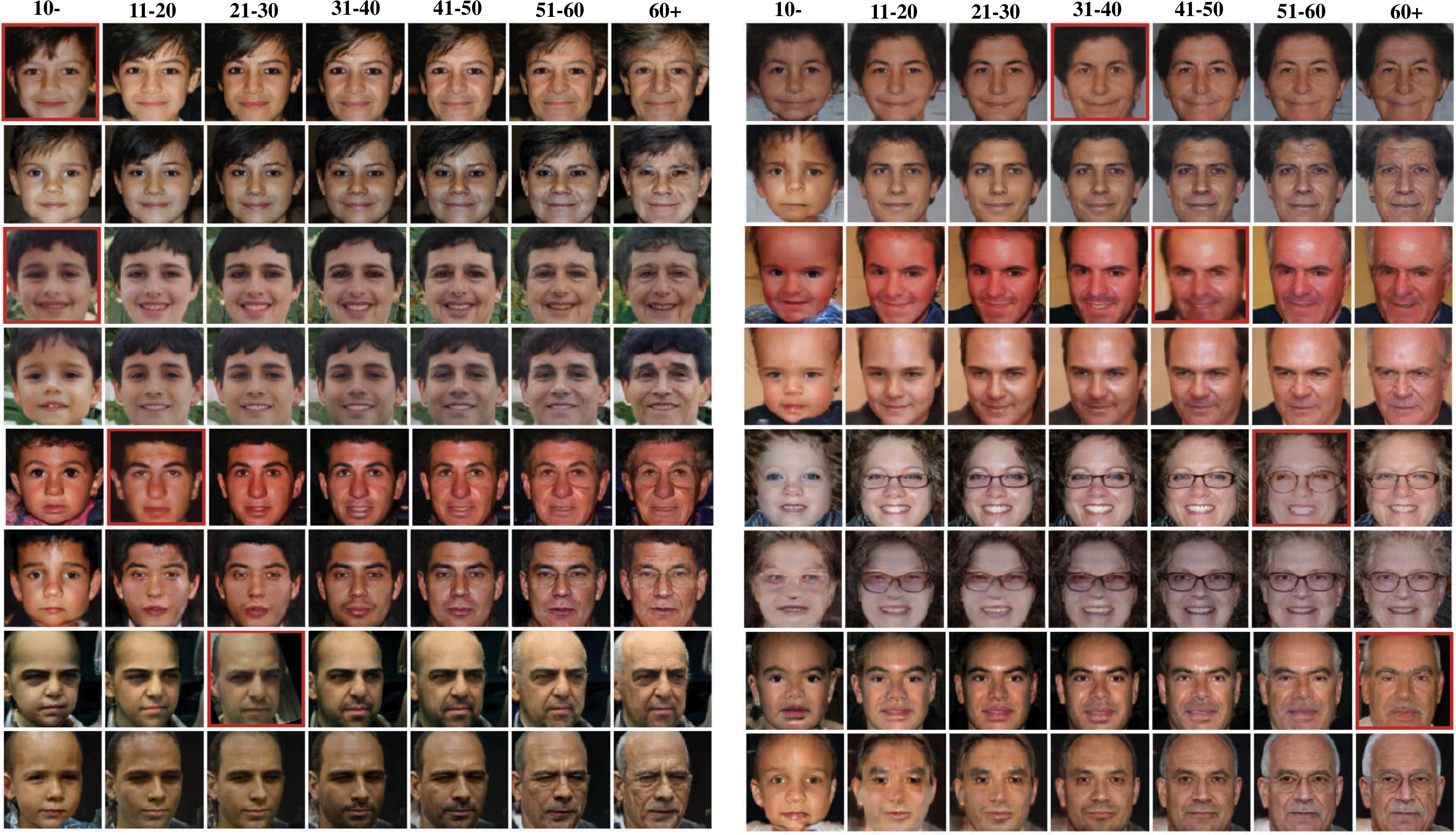}
    \caption{Qualitative comparisons between \methodname with and without AIFR.  The first and second rows of each face show the results of \methodname with and without AIFR, respectively, where the input images are highlighted in red boxes. Obviously, without AIFR, the synthesized faces cannot preserve the input identities and produce photorealistic aging/rejuvenated effects.
    }
    \label{fig:qualitative_comparison_wo_AIFR}
\end{figure*}

\begin{figure*}[h]
    \centering
    \includegraphics[width=0.7\linewidth]{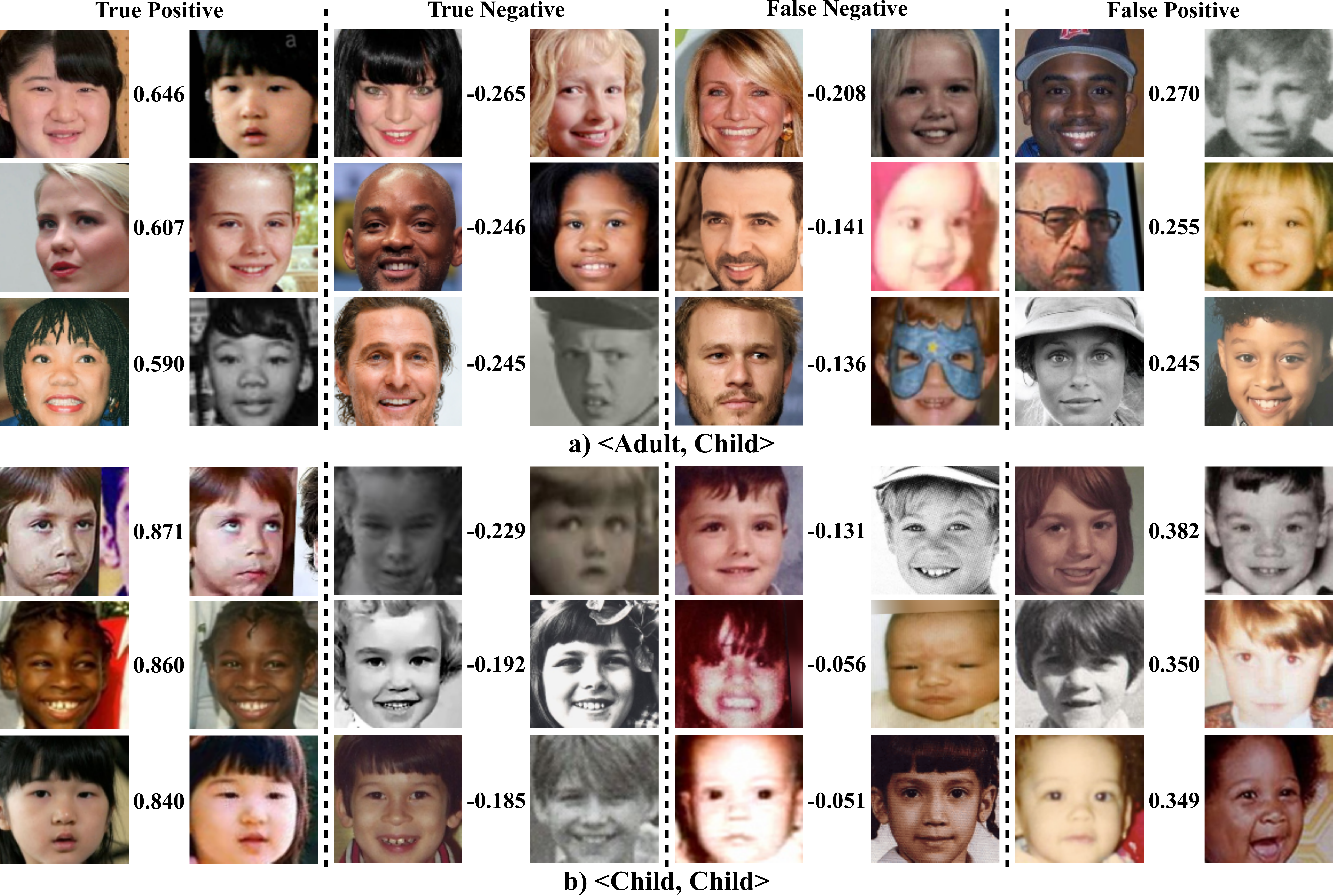}
    \caption{Example pairs on \testsetname for two sub-tasks.
    After computing and sorting the cosine similarity for all pairs of two tasks, we showcase the true positives and false positives with the highest similarity and the true negatives and false negatives with the lowest similarity. The results suggest that the large age gaps of \testsetname have a significant impact on the face recognition models, which deserves more research attention based on \testsetname for the AIFR task.
    }
    \label{fig:case_analysis}
\end{figure*}

This supplementary material provides the following extra contents: (1) Sec.~\ref{sec:comparisons_to_conf} presents the comparisons to our preliminary version~\cite{huang2021age} including qualitative comparison in Fig~\ref{fig:qualitative_comparison_r1} and quantitative comparison in Table~\ref{tab:comparisons_to_conf}; (2) Sec.~\ref{sec:clarify_diff} clarifies the differences between StyleGAN and our \methodname for continuous synthesis, summarizes the advantages of \methodname, and conducts extensive quantitative and qualitative comparisons; and
(3) Sec.~\ref{sec:additional_experiments} provides the comparisons between \methodname with and without AIFR in Fig.~\ref{fig:qualitative_comparison_wo_AIFR}, and example pairs on ECAF for two sub-tasks in Fig.~\ref{fig:case_analysis}.

\section{Comparisons to Our Preliminary Version~\cite{huang2021age}}
\label{sec:comparisons_to_conf}

\begin{figure*}[h]
    \centering
    \includegraphics[width=0.95\linewidth]{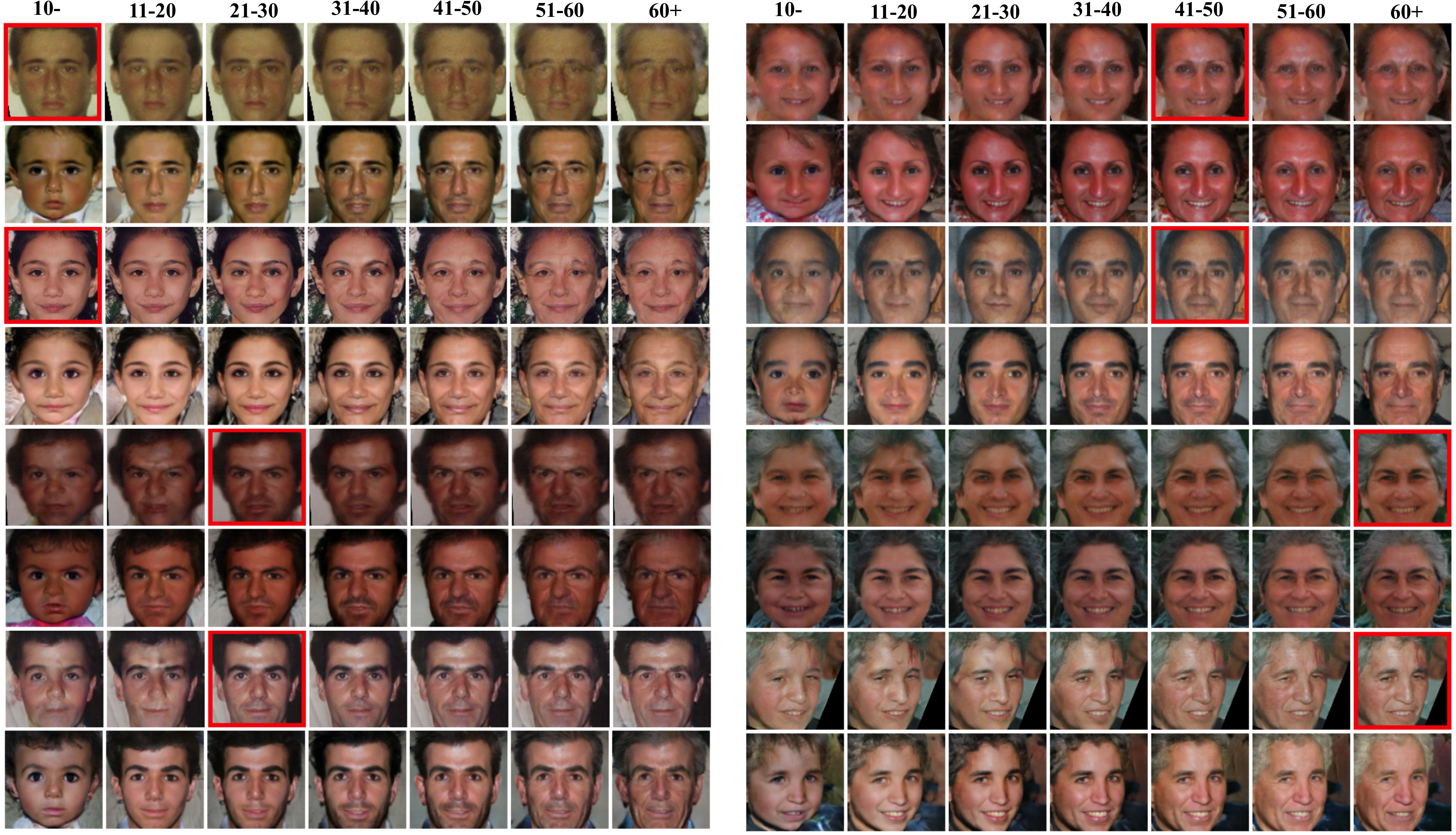}
    \caption{
        Comparisons to the preliminary version~\cite{huang2021age} for FAS task on FG-NET. The first and second rows of each face show the results of the preliminary version and the new model, respectively, with the input images highlighted by red boxes. Our new model has significantly improved the quality of synthesized faces in terms of various poses, gender, etc with much more natural aging/rejuvenated effects; \eg, the white hair and beard grow during face aging.
        We note that the background can be preserved by extracting the face area from the input images prior to being fed into the synthesis network, as discussed in Sec.~5.
    }
    \label{fig:qualitative_comparison_r1}
\end{figure*}

In this section, we present the comparisons between the preliminary version~\cite{huang2021age} and the new \methodname,  qualitatively for FAS in Fig.~\ref{fig:qualitative_comparison_r1} and quantitatively for AIFR in Table~\ref{tab:comparisons_to_conf}.

\begin{table}[h]
    \centering
    \caption{Comparisons to the preliminary version~\cite{huang2021age} on several benchmarked datasets for AIFR task. Our new model has consistently outperformed the preliminary version~\cite{huang2021age} with the newly proposed \seldaname.
    \seldaname achieves 0.60 and 0.50 performance improvements on FG-NET~(MF1) and ECAF~$<$Adult, Child$>$, respectively. Although there are marginal improvements on LFW and CACD-VS, we think that the current results have almost reached the best on these small datasets (very close to 100\%), which is hard to make more improvements.
    }
    \begin{tabular}{lcc}
    \toprule
    Dataset  &  \methodname~\cite{huang2021age}  & \methodname \\
    \midrule
    AgeDB-30  & 96.23 & \textbf{96.45} ($\uparrow$ 0.22) \\
    CALFW  & 95.62 & \textbf{95.98} ($\uparrow$ 0.36) \\
    CACD-VS  & 99.55 & \textbf{99.58} ($\uparrow$ 0.03) \\
    FG-NET~(leave-one-out) & 94.78 & \textbf{95.00} ($\uparrow$ 0.22)\\
    FG-NET~(MF1)  & 57.18 & \textbf{57.78} ($\uparrow$ 0.60)\\
    ECAF~$<$Child, Child$>$  & 91.05 & \textbf{91.20} ($\uparrow$ 0.15)\\
    ECAF~$<$Adult, Child$>$   & 87.05 & \textbf{87.55} ($\uparrow$ 0.50)\\
    LFW  & 99.52 & \textbf{99.55} ($\uparrow$ 0.03)\\
    MF1-Facescrub  & 77.06 & \textbf{77.33} ($\uparrow$ 0.27)\\
    \bottomrule
    \end{tabular}
    \label{tab:comparisons_to_conf}
\end{table}

\newpage
\section{Relation to StyleGAN}
\label{sec:clarify_diff}
Here, we first summarize \emph{how to achieve FAS with StyleGAN in previous work~\cite{karras2019style,karras2020analyzing} and in our \methodname, and then discuss their differences}.

\noindent\textbf{Previous work leveraging StyleGAN for FAS.}\quad
In previous work, the decoder $D$ of StyleGAN learns to generate images $D(\vct{z})$ from the latent variable $\vct{z}\in \mathcal{Z}$, which is sampled from a simple prior distribution such as standard Gaussian distribution. Given a facial image $\mat{I}$, they need three steps to achieve the face age synthesis task.
\begin{enumerate}
\item The image $\mat{I}$ is first inversed into the latent space $\mathcal{Z}$, which is usually achieved by optimizing the following loss function:
\begin{align}
    \vct{z}^{*}=\underset{\vct{z}}{\arg \min }\ \ell(D(\vct{z}), \mat{I}),\label{eq:gan_inversion}
\end{align}
where $l$ can be the perceptual or pixel-wise loss functions to encourage the consistency between input and reconstructed faces. 

\item The semantic manipulation direction is computed from a set of images with accurate age labels:
\begin{align}
    \vct{n}_{s\to t}=\vct{\mu}_t - \vct{\mu}_s,
\end{align}
where $\vct{\mu}_i=\mathrm{mean}(\{\vct{z}\}_{y_{\mathrm{age}}=i})$ is the mean cluster centers of each age group, $s$ and $t$ are the input and target age groups, respectively.

\item The latent code is manipulated along the semantic direction to obtain the latent code at the target age group:
\begin{align}
    \vct{\widehat{z}}_t=\vct{z}^{*} + \vct{n}_{s\to t},\label{eq:gan_manipulation}
\end{align}
which is then decoded to get the visual result $\mat{\widehat{I}}_t=D(\vct{\widehat{z}}_t)$. As a byproduct, StyleGAN achieves continuous face age synthesis by interpolating between two adjacent age groups $\alpha \vct{\widehat{z}}_t + (1-\alpha) \vct{\widehat{z}}_{(t+1)}$, where $\alpha \in [0, 1]$ is the coefficient for interpolation.
\end{enumerate}

\noindent\textbf{Our \methodname leveraging StyleGAN for FAS.}\quad
In \methodname, the process of rendering the input face $\mat{I}$ to the synthesized face $\mat{\widehat{I}}_t$ that belongs to the target age group $t$ is:
\begin{align}
    \mat{C}_t^1 &= f_1(\mat{X}_{\mathrm{id}}, \mat{E}^1, t), \\
    \mat{C}_t^l &= f_l(\mat{C}_t^l, \mat{E}^l, t), \quad l\in \{2,3\} \\
    \mat{\widehat{I}}_t &= D(\mat{X}_{\mathrm{id}}, \{\mat{C}_t^l\}_{l=1}^{3}),\label{eq:aging}
\end{align}
where $l$ denotes the index of different levels, and $\mat{C}_l$ and $f_l$ are the identity-level age condition and ICMs at the $l$-th level, respectively.
Similarly, we can obtain the intermediate faces $D(\mat{X}_{\mathrm{id}}, \{\alpha \mat{C}_t^l + (1-\alpha) \mat{C}_{(t+1)}^l \}_{l=1}^{3})$ by interpolating between adjacent $\mat{C}_l$ to achieve continuous face age synthesis.

\begin{figure*}[t]
    \centering
    \includegraphics[width=0.8\linewidth]{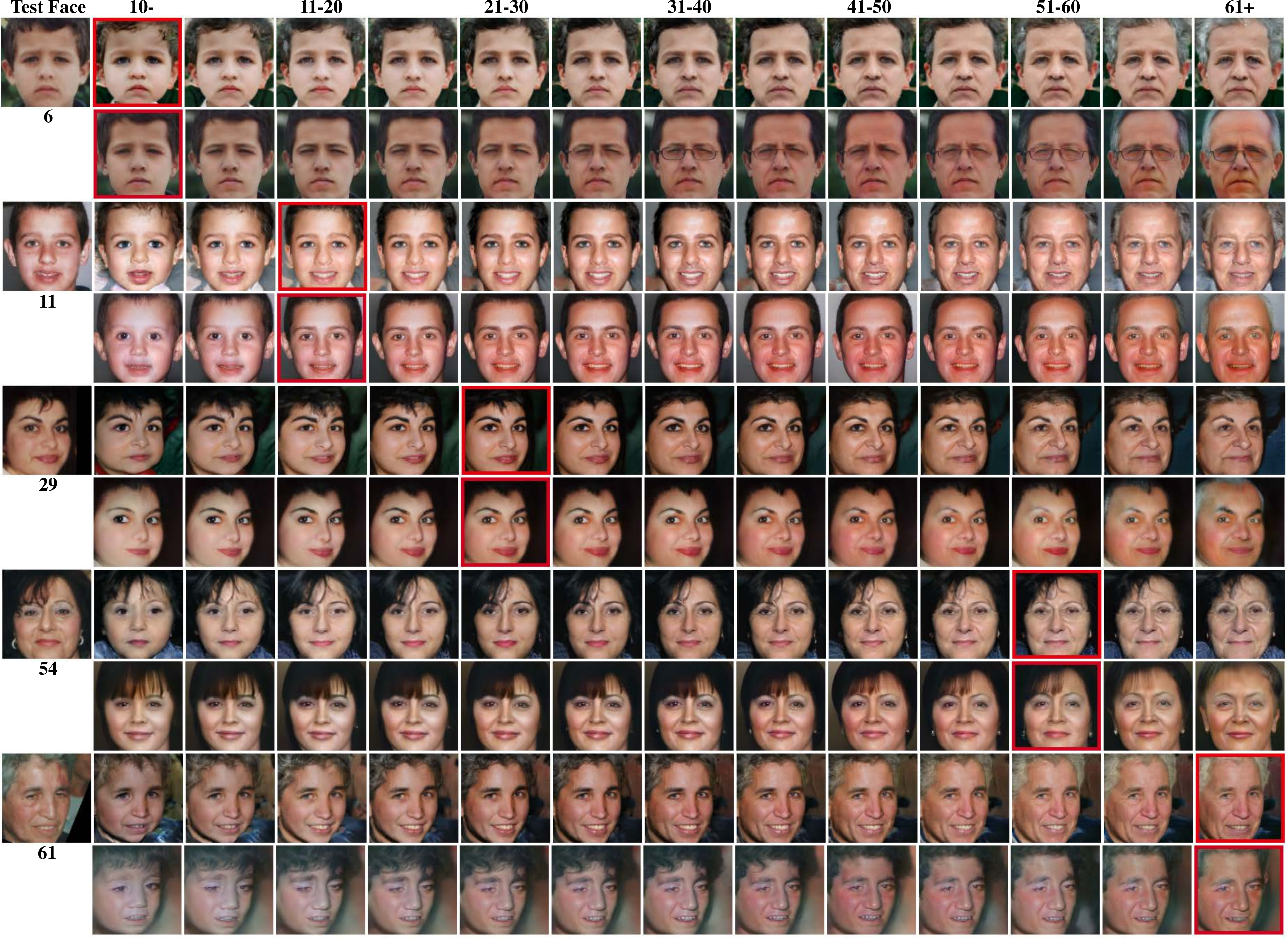}
    \caption{Sample results of continuous face age progression and regression for \methodname~(first row) and StyleGAN2~(second row) on the FG-NET dataset. We show the test faces in the first column, with the age below the image. The red boxes indicate the reconstructed faces in the same age groups as the input faces. We synthesize the continuous aging/rejuvenation faces by interpolating between the two latent variables of adjacent age groups. Zoom in for a better view of image details.
    We note that the background can be preserved by extracting the face area from the input images prior to being fed into the synthesis network, as discussed in Sec.~5.
    }
    \label{fig:continuous_face_age_synthesis_a}
\end{figure*}

\noindent\textbf{Advantages of \methodname over previous work.}\quad
Compared to simply interpolating between the latent variables in StyleGAN, the advantages for the proposed \methodname can be summarized as follows:
\begin{enumerate}
    \item \emph{Real-time face age synthesis}. The computation cost of StyleGAN is expensive as the optimization process in Eq.~\eqref{eq:gan_inversion} takes a lot of time for searching the optimal latent variable. Table~\ref{tab:comparisons} shows the time to process a single image, where StyleGAN takes longer time than other methods~(we have used the official code on the same image size). On the contrary, \methodname can perform real-time face age synthesis while achieving pleasing synthesis results.
    \item \emph{No requirements for age labels during inference}. 
    Compared to \methodname, StyleGAN does not require age labels during training. However,
    during inference, StyleGAN requires the age labels to obtain the corresponding semantic direction as shown in the manipulation process in Eq.~\eqref{eq:gan_manipulation}, while \methodname can synthesize faces at any age without the age labels. Therefore, \methodname is more flexible than directly interpolating in the latent space. 
    \item \emph{More natural aging/rejuvenated effects}. Fig.~\ref{fig:continuous_face_age_synthesis_a} shows the comparisons of \methodname and StyleGAN. Although both of them produce smooth aging/rejuvenated results, \methodname can render faces with more natural aging/rejuvenated effects. For example, \methodname has changed the head shape for child's faces at 10- age group, while StyleGAN only simply whitens the skin.
\end{enumerate}

\noindent\textbf{Effectiveness of shared feature.}\quad
According to Eq.~\eqref{eq:aging}, the difference between vanilla StyleGAN and \methodname is that there is multi-level $f_l$ receiving $\mat{E}^l$ and $t$ to generate the condition $\mat{C}_l$. Therefore, we can evaluate the effectiveness of the shared features, or $f_l$, by the quantitative comparisons, which are reported in Table~\ref{tab:comparisons}. There are two observations:
\begin{enumerate}
    \item Vanilla StyleGAN can \emph{not} preserve the identities of input faces very well~(see lower cosine similarity). We would like to emphasize that the ages are highly correlated to other facial attributes such as genders and races in the latent space, which may be changed when interpolating along the manipulation directions.
    \item Vanilla StyleGAN does \emph{not} produce the faces at the target age group~(see lower age accuracy). The manipulation directions are shared across different persons while different persons may have different aging/rejuvenation trends.
\end{enumerate}

The above problems have been solved by the proposed multi-level ICMs. The input identities can be well preserved by $\mat{X}_{\mathrm{id}}$ and the shared features can enforce the model  to produce smooth age conditions.

\begin{table}[t]
    \caption{
        Quantitative comparisons on FG-NET and the inference time for different methods~(sec/img). The quantitative results are in the form of $a/b/c$, where $a$, $b$, and $c$ represent the mean values of age accuracy~(\%), the mean absolute error between the predicted ages of generated and real faces, and identity preservation (cosine similarity) computed over all age mappings, respectively. The official code and pre-trained model of StyleGAN~\cite{karras2020analyzing} have been used to report the quantitative results. We have only reported the results on FG-NET, since the inversion process of StyleGAN takes too much time. The inference time is measured on one single NVIDIA A100 GPU and the same image size of 128.
    }
    \label{tab:comparisons}
    \centering
    \begin{tabular}{lcr}
    \toprule
    Method          & Quantitative comparisons  & Speed~(sec/img)\\ 
    \midrule
    CAAE~\cite{zhang2017age}  & 40.72/9.32/0.146\std{0.097} & 0.0005 \\ 
    IPCGAN~\cite{wang2018face}  & 61.22/4.32/0.453\std{0.122} & 0.0034 \\
    S2GAN~\cite{he2019s2gan}  & 60.26/4.37/0.293\std{0.111} & 0.0011 \\ 
    StyleGAN~\cite{karras2020analyzing}  & 45.55/7.64/0.260\std{0.131} & 46.3000 \\ 
    MTLFace (\textbf{ours}) & \textbf{71.33}/\textbf{3.10}/\textbf{0.648}\std{0.094} & 0.0095 \\ 
    \bottomrule
    \end{tabular}
\end{table}

\clearpage

\section{Additional Experiments}
\label{sec:additional_experiments}

\end{document}